\documentclass{article}

\PassOptionsToPackage{numbers, compress}{natbib} %
\usepackage[final, main]{neurips_2026} %
\usepackage[dvipsnames]{xcolor}
\usepackage[hidelinks]{hyperref} %
\usepackage{amsmath}
\usepackage{cleveref}
\usepackage{url}

\usepackage{enumitem}

\crefname{figure}{Fig.}{Figs.}
\Crefname{figure}{Figure}{Figures}
\crefname{section}{Sec.}{Secs.}
\Crefname{section}{Section}{Sections}
\crefname{table}{Tab.}{Tabs.}
\Crefname{table}{Table}{Tables}

\usepackage{caption}
\usepackage{booktabs}
\usepackage{array}
\newcolumntype{M}{>{\centering\arraybackslash}m{3.2em}}
\newcolumntype{L}{>{\centering\arraybackslash}m{4.2em}}
\newcolumntype{G}{>{\centering\arraybackslash}m{5.2em}}

\usepackage[ruled,vlined]{algorithm2e}
\usepackage[most]{tcolorbox}
\usepackage[table]{xcolor}
\definecolor{best}{RGB}{220,240,220}
\definecolor{second}{RGB}{228,243,228}
\tcbuselibrary{listingsutf8}
\usepackage{wrapfig}

\newtcolorbox{inlinealgo}[1][]{
  colback=gray!5,
  colframe=black!60,
  fonttitle=\bfseries,
  title=Self-Correction Sampling,
  boxrule=0.4pt,
  arc=2pt,
  left=6pt,
  right=6pt,
  top=4pt,
  bottom=4pt,
  enhanced,
  width=\linewidth,
  #1
}

\usepackage[normalem]{ulem}

\def \customparskip {.3em}
\renewcommand{\paragraph}[1]{\vspace{\customparskip}\noindent\textbf{#1}}

\definecolor{turquoise}{cmyk}{0.65,0,0.1,0.3}
\definecolor{purple}{rgb}{0.65,0,0.65}
\definecolor{dark_green}{rgb}{0, 0.5, 0}
\definecolor{orange}{rgb}{0.8, 0.6, 0.2}
\definecolor{red}{rgb}{0.8, 0.2, 0.2}
\definecolor{darkred}{rgb}{0.6, 0.1, 0.05}
\definecolor{blueish}{rgb}{0.0, 0.3, .6}
\definecolor{light_gray}{rgb}{0.7, 0.7, .7}
\definecolor{pink}{rgb}{0.9, 0, 0.6}
\definecolor{greyblue}{rgb}{0.25, 0.25, 1}
\definecolor{teal}{rgb}{0.0, 0.4, 0.4}

\newcommand{\jg}[1]{{\color{greyblue}#1}}

\usepackage[table]{xcolor}

\renewcommand{\paragraph}[1]{\smallskip\noindent\textbf{#1}}

\usepackage[utf8]{inputenc} %
\usepackage[T1]{fontenc}    %
\usepackage{hyperref}       %
\usepackage{url}            %
\usepackage{booktabs}       %
\usepackage{amsfonts}       %
\usepackage{nicefrac}       %
\usepackage{microtype}      %
\usepackage{xcolor}         %
\usepackage{multirow}
\usepackage{wrapfig}
\usepackage{lipsum}

\title{DDMS: \underline{D}iscriminative \underline{D}istillation of \underline{M}ulti-view Foundational Features into \underline{S}ingle-view Models}

\author{
Jeong-gi Kwak$^{1}$ \quad Sho Kagami$^{2}$ \quad Yuki Ono$^{3}$ \quad Kwang Moo Yi$^{1}$\\[0.3cm]
$^{1}$University of British Columbia\\
$^{2}$Sony Semiconductor Solutions Corporation\\
$^{3}$Sony Corporation
}

\begin{document}

\maketitle

\begin{abstract}
Foundational visual features such as DINO have played a critical role across modern computer vision, and have recently become key components in multi-view feed-forward geometry estimators.
In this work, we demonstrate that by re-distilling these multi-view models---their internal knowledge of 3D geometry---into a single-view estimator, we can obtain enhanced 3D consistent foundational features.
Our key idea is to construct a \emph{multi-view teacher} by fusing pretrained 2D foundation features with multi-view geometric features, and refining the fused representation with a discriminative ranking objective.
Through our discriminative distillation framework, we enforce the learned features to be \emph{both 3D consistent and locally distinctive}, while keeping them aligned with the feature space of the original foundation model to preserve the semantic structure of the pretrained representation.
Consistency and local discriminability are critical for 3D computer vision problems such as forming semantic and geometric correspondences across images.
To demonstrate the effectiveness of our method, we perform comprehensive experiments spanning multiple angles: direct feature analysis, dense prediction transfer, and explicit 3D lifting and rendering.
Across these evaluations, our method consistently produces stronger 3D-aware foundation features that improve multi-view consistency and local discriminability while preserving the semantic transferability of the original representation.

\end{abstract}

\section{Introduction}
\label{sec:intro}

Recent vision foundation models~\cite{caron2021dino,radford2021learning_clip,oquab2023dinov2,simeoni2025dinov3} pretrained on large-scale image data have emerged as powerful general-purpose visual backbones.
Their features exhibit strong semantic transferability~\cite{amir2021deepvit,hamilton2022stego,dong2023maskclip} and support a wide range of recognition and dense prediction tasks.
However, because these features are primarily learned from single-image or image-level pretraining objectives~\cite{caron2021dino,he2022masked,oquab2023dinov2}, they often exhibit limited 3D awareness under multi-view evaluation~\cite{el2024probing}.
In particular, features extracted from different views of the same scene or object may vary across viewpoints, while visually similar but geometrically distinct regions can remain insufficiently separated in feature space.
Such inconsistency makes it difficult to reliably associate or aggregate features across viewpoints, limiting their effectiveness in 3D vision tasks that depend on stable cross-view feature associations.

\begin{figure*}[t]
\centering
\includegraphics[width=\linewidth]{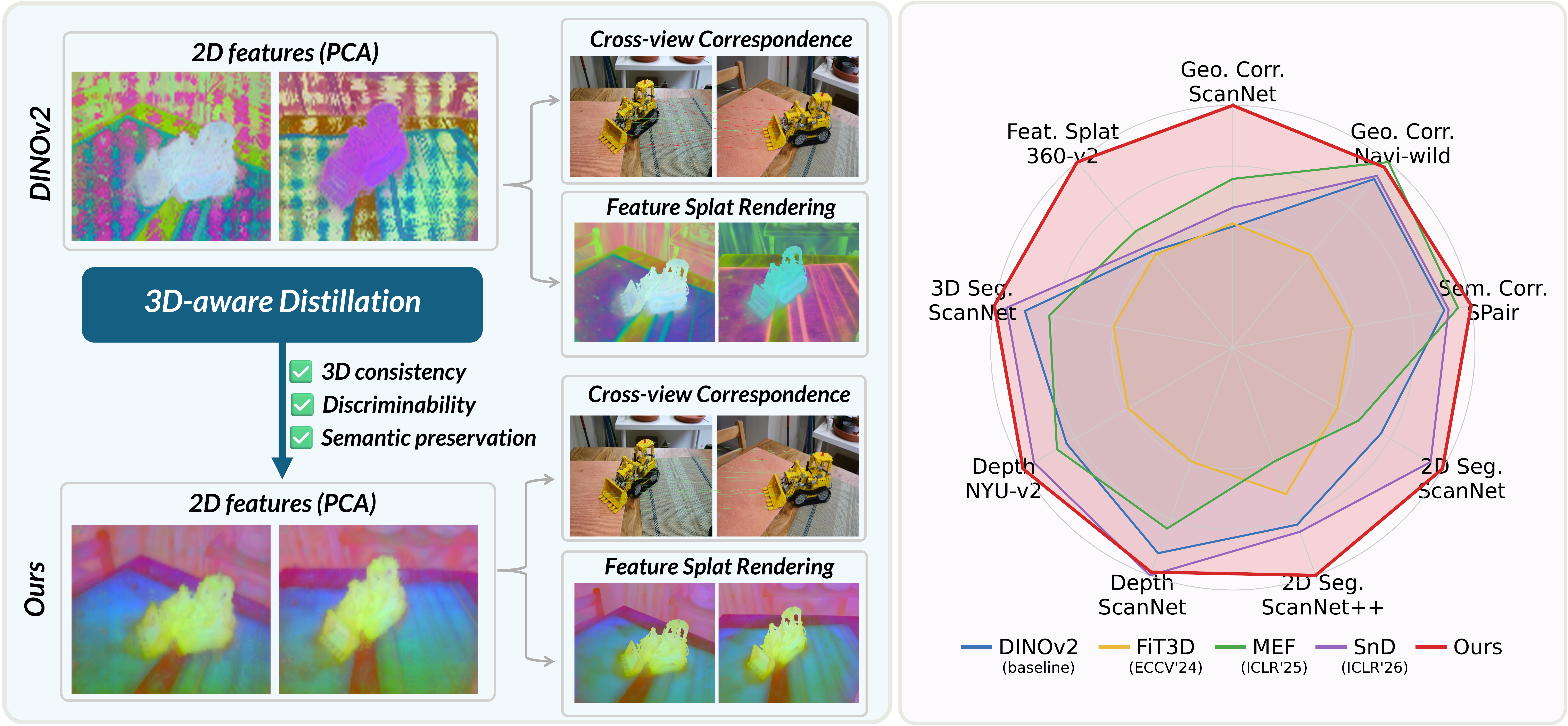}
\captionof{figure}{
Our 3D-aware distillation turns pretrained 2D foundation features into representations that are more consistent across views while retaining their semantic utility.
The resulting features produce cleaner cross-view correspondences and remain coherent after explicit 3D feature splatting (left).
Across 3D-aware finetuning and distillation baselines~\cite{fit3d,mef,snd} built from the same 2D foundation backbone~\cite{oquab2023dinov2}, our representation provides a more balanced profile over geometric correspondence, semantic transfer, dense prediction, and 3D lifting evaluations (right).
The radar chart axes use per-metric min--max normalization with the minimum set to 50\%.}
\label{fig:intro}
\end{figure*}
To address this limitation, recent works~\cite{fit3d,mef,snd} have sought to make pretrained visual features more consistent across views and more 3D-aware.
A prominent direction uses 3D Gaussian Splatting (3DGS)~\cite{kerbl20233dgs} as an intermediate representation to construct multi-view feature supervision.
Fit3D~\cite{fit3d} lifts pretrained 2D foundation features into scene-specific 3D Gaussian representations and renders them from multiple views to supervise backbone fine-tuning.
SnD~\cite{snd} improves the scalability of this idea by replacing scene-specific optimization with a feed-forward 3DGS pipeline~\cite{chen2024mvsplat}, distilling rendered 3D-aware features back into the backbone.
These methods, however, encourage features from different views to become similar, which can smooth out fine-grained distinctions that are critical for image matching, a foundational building block of 3D computer vision~\cite{lowe2004sift,schonberger2016sfm,sarlin2020superglue}.
As shown in \Cref{fig:intro}, while the 3D consistency of semantic segmentation improves, geometric and semantic correspondence performances can be suboptimal, or even worse than the base foundational feature, as shown in the case of Fit3D~\cite{fit3d}.

Naturally, another line of work aims to improve feature discriminability, more in line with traditional correspondence-based descriptor learning~\cite{mishchuk2017hardnet,luo2018geodesc}.
MEF~\cite{mef} tunes pretrained features to serve as stronger matching descriptors for multi-view correspondence.
Again, while this improves correspondence performance across views, it slightly weakens the semantic transferability of original foundational features as shown in \Cref{fig:intro}, making them more suitable as image correspondence features but not as general foundational features.

In this work, instead, we turn our attention to multi-view foundational models~\cite{wang2024dust3r,wang2025vggt,lin2025da3} that have recently shown impressive performance.
Although these models are trained for geometric prediction rather than general-purpose feature extraction, their internal representations have been shown to encode correspondence and epipolar geometric cues~\cite{bratuli2026geometric}, suggesting that they capture visual information useful for 3D reasoning.
We therefore use these features together with foundational 2D features~\cite{caron2021dino,oquab2023dinov2} to derive a new representation that carries the best of both: the semantic utility of 2D foundation features and the geometric awareness of multi-view models.
Then, to maximize utility at inference time, we distill the resulting \emph{multi-view teacher} into a \emph{single-view student}.

When learning the \emph{multi-view teacher}, our goal is to jointly promote 3D consistency, fine-grained separability, and semantic preservation.
To this end, we leverage the rich geometric priors of the multi-view model through a ranking-based objective, while anchoring the refined features to the original 2D foundation feature space.
The ranking objective encourages features from the same local 3D surface to rank above geometrically incompatible candidates, rather than merely increasing similarity among all cross-view observations.
At the same time, feature-space anchoring keeps the refinement aligned with the manifold of the original feature space (\textit{e.g.}, DINO~\cite{caron2021dino,oquab2023dinov2}), helping preserve the semantic structure of the pretrained representation.
After distillation, the resulting single-view features become more 3D-consistent, yield cleaner cross-view correspondences, and remain coherent after 3D feature lifting, as illustrated in \cref{fig:intro}.
Through comprehensive experiments across multiple complementary evaluations in \cref{sec:result}, we further show that these gains are not confined to a single aspect of the representation.
Our method maintains strong performance across geometric, semantic, and 3D aggregation settings, demonstrating a favorable balance between cross-view consistency, discriminability, and semantic preservation.

We summarize our main contributions as follows:
\begin{itemize}[leftmargin=*]
    \item we introduce a 3D-aware geometric foundational feature distillation framework that utilizes a multi-view teacher and a single-view student;
    \item to improve cross-view consistency while preserving separability and semantic structure, we propose to train the multi-view teacher via
    geometry-supervised ranking and feature-space anchoring; and
    \item %
    with our single-view student, we
    demonstrate consistent gains across correspondence, dense prediction transfer, and explicit 3D lifting evaluations.
\end{itemize}

\section{Related Work}
\label{sec:related}

\paragraph{Visual foundation models and 3D awareness.}
Recent visual foundation models~\cite{caron2021dino,oquab2023dinov2,darcet2023vision,zhou2021ibot,he2022masked,radford2021learning_clip,touvron2022deit,ranzinger2024radio,sariyildiz2025dune,simeoni2025dinov3,kirillov2023segment} have shown strong transfer across recognition, dense prediction, segmentation, and correspondence-based tasks.
A key reason for their broad utility is that their intermediate features often behave as dense visual representations rather than only image-level descriptors.
This has enabled their use as dense visual descriptors~\cite{amir2021deepvit}, for unsupervised semantic grouping and segmentation~\cite{hamilton2022stego,dong2023maskclip}, as general-purpose dense features~\cite{oquab2023dinov2,darcet2023vision,fu2024featup}, and for point tracking or temporal correspondence~\cite{tumanyan2024dinotracker}.
Beyond 2D recognition and dense prediction, such features have also become important components in 3D vision pipelines, including dense image matching and correspondence estimation~\cite{edstedt2024roma,sarlin2020superglue,sun2021loftr,lindenberger2023lightglue,edstedt2025romav2}, visual odometry and SLAM~\cite{mur2015orb,teed2021droid,azhari2025dinovo}, feature-based 3D lifting and feature field construction~\cite{tschernezki2022n3f,zhou2024feature3dgs,marrie2025ludvig}, and open-vocabulary or language-aligned 3D scene understanding~\cite{peng2023openscene,kerr2023lerf,qin2024langsplat}.
These applications all rely, either explicitly or implicitly, on stable feature associations across views.
However, generic visual foundation features are mostly learned from single images or image-level supervision and are not explicitly optimized for cross-view geometric consistency.
Prior probing work~\cite{el2024probing} shows that such features contain useful but limited 3D awareness, with performance becoming less reliable under large viewpoint changes and multi-view correspondence settings.
This limitation motivates recent efforts to adapt pretrained visual features with geometric, multi-view, or distillation-based supervision.

\paragraph{3D-aware feature learning and distillation.}
To address the limited 3D awareness of pretrained visual features, recent works have explored geometric supervision, 3D lifting, and distillation-based adaptation.
Prior work has used multi-view or RGB-D geometry for representation learning~\cite{prid3d,3dgstfm,croco,crocov2}; in contrast, DDMS adapts a pretrained 2D foundation model by constructing a teacher from frozen multi-view geometry features and distilling it into a single-view encoder.
DUNE~\cite{sariyildiz2025dune} studies a broader form of feature distillation by combining heterogeneous 2D and 3D teachers into a universal visual encoder.
More directly related to our setting, several methods adapt a pretrained DINOv2 backbone~\cite{oquab2023dinov2} with explicit multi-view or 3D supervision.
Fit3D~\cite{fit3d} lifts 2D foundation features into scene-specific 3D Gaussian representations and renders them from novel views to fine-tune the backbone.
SnD~\cite{snd} improves scalability by replacing per-scene feature optimization with a feed-forward 3DGS pipeline, and incorporates SAM~\cite{kirillov2023segment} masks for mask-aware feature upsampling and blending to strengthen dense semantic transfer.
While effective for enforcing cross-view agreement, rendering- and aggregation-based supervision can smooth fine-grained feature differences, limiting local discriminability.
MEF~\cite{mef} takes a complementary direction by adapting pretrained features toward multi-view matching descriptors.
This improves correspondence performance, but the resulting descriptor-oriented representation can become less suitable as a general-purpose dense feature.

\paragraph{Multi-view geometry and feed-forward 3D models.}
Recent multi-view geometry foundation models~\cite{wang2024dust3r,wang2025vggt,lin2025da3,wang2025pi,keetha2025mapanything} have shifted 3D reconstruction from per-scene optimization toward generalizable feed-forward prediction.
Given one or more images, these models predict geometric quantities such as depth, point maps, camera parameters, tracks, or metric scene structure using strong image encoders and cross-view reasoning.
Feed-forward 3DGS methods~\cite{chen2024mvsplat,xu2025depthsplat,huang2025noposplat} similarly produce renderable 3D representations without per-scene optimization.
Beyond geometry prediction, 2D foundation features have also been lifted or aggregated into 3D representations for semantic and open-vocabulary scene understanding~\cite{sun2025uni3r,li2025semanticsplat,ye2026semgs}.
This feed-forward 3D paradigm has also been adopted for 3D-aware feature distillation.
SnD~\cite{snd}, for example, renders feature supervision from a feed-forward 3DGS pipeline~\cite{chen2024mvsplat} to distill a 2D encoder.
Rather than relying solely on rendered feature supervision, we use the intermediate features of a frozen multi-view geometry foundation model as multi-view geometric context for teacher construction.
This choice is motivated by recent analyses showing that feed-forward 3D reconstruction models encode correspondence and epipolar cues in their intermediate representations~\cite{bratuli2026geometric}.
This allows the teacher to combine geometric priors with the semantic structure of pretrained 2D features before distillation.
\section{Method}
\label{sec:method}

\begin{figure*}[t]
\centering
\includegraphics[width=\linewidth]{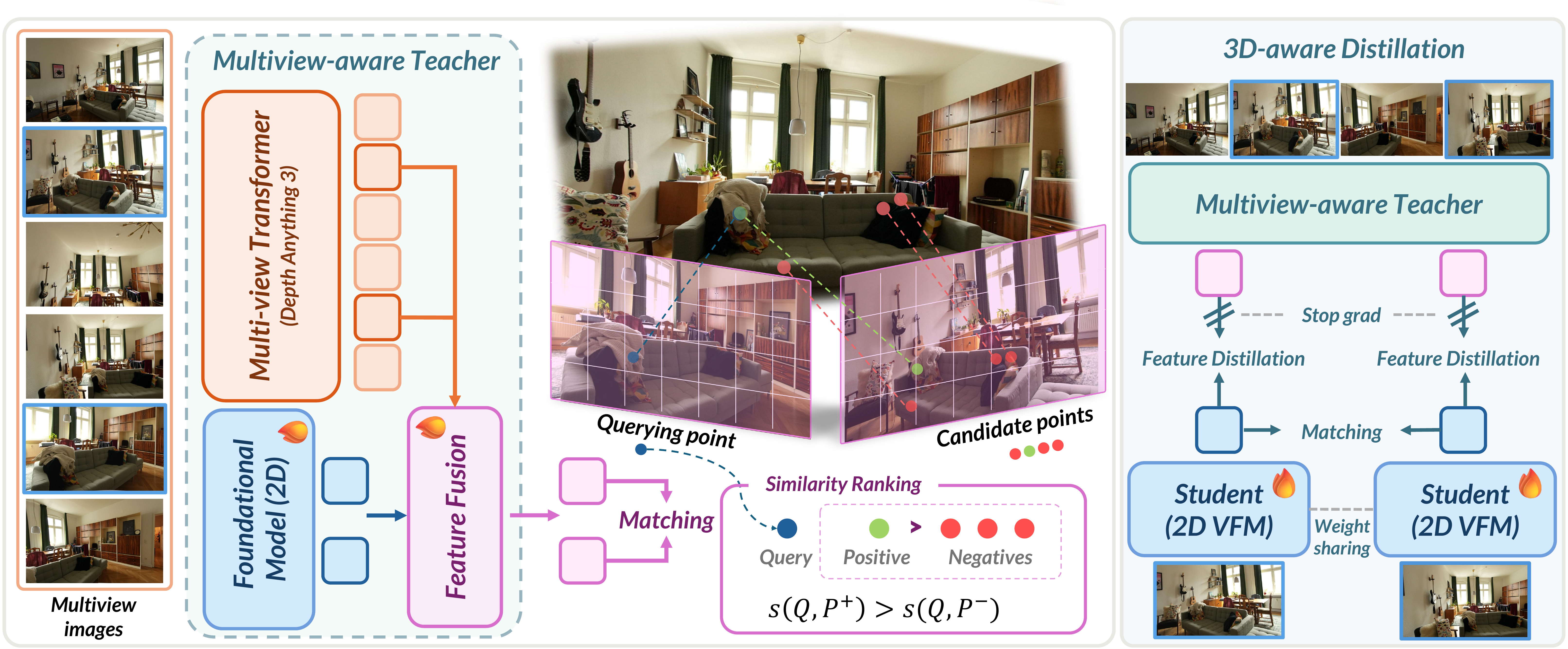}
\captionof{figure}{
Overview of our method.
We fuse pretrained 2D foundation features~\cite{oquab2023dinov2} with frozen multi-view geometry features~\cite{lin2025da3} to construct a multi-view teacher.
We refine the teacher with geometry-supervised ranking and feature-space anchoring, encouraging valid 3D correspondences to rank above incompatible candidates while preserving the original semantic feature space.
We then distill it into a single-view student, which is used as the final feature extractor at inference time.
}
\label{fig:method}
\end{figure*}

\jg{}
Our goal is to adapt pretrained visual features into 3D-aware representations that improve cross-view consistency without sacrificing discriminability or the semantic richness of the original representation.
To this end, we first construct a multi-view-aware teacher by enriching a 2D visual foundation model with geometric multi-view priors (\cref{sec: teacher_construction}), and further refine it through discriminative learning (\cref{sec:feature_learning}). We then distill the resulting teacher into a student visual encoder (\cref{sec:distillation}). The key idea is to inject multi-view geometric awareness into feature learning while maintaining discriminative matching ability and the semantic prior of the original 2D representation.

\subsection{Overview} \label{sec: overview}
As illustrated in \cref{fig:method}, our framework consists of two stages.
First, we construct a multi-view-aware teacher by fusing features from a pretrained 2D visual foundation model with geometry-aware features from a multi-view transformer.
The teacher is refined with a geometry-guided ranking objective, which ranks geometrically corresponding points above spatially distant 3D points in feature space.
Second, we distill this teacher into a single-view student encoder using feature-level supervision.
This allows the final student to be used like a standard 2D backbone at inference time, while inheriting multi-view geometric awareness from the teacher.

\subsection{Multi-view-aware Teacher Construction} \label{sec: teacher_construction}

Given a set of $N$ input views $\mathcal{I}=\{I_i\}_{i=1}^{N}$, we first run a geometric foundation model, implemented as a multi-view transformer, on all views jointly.
Through cross-view attention, the model produces geometry-aware features whose representation for each view is contextualized by the full input set, allowing it to encode spatial relationships beyond what can be inferred from a single image alone.
Feature learning is performed on a subset of views $V \subset \{1,\ldots,N\}$ with $|V|=M$.
For each selected view $m \in V$, a pretrained 2D visual foundation model processes $I_m$ independently and produces semantic features $F^v_m$, while we extract an intermediate feature $F^g_m$ from the multi-view transformer~\cite{lin2025da3}, where correspondence and epipolar cues are known to emerge~\cite{bratuli2026geometric}.
Since $F^g_m$ is optimized for geometric prediction rather than general-purpose visual representation, we leverage it as geometric context for refining the semantically rich 2D feature $F^v_m$.
The two representations are concatenated on the same ViT patch grid.
A lightweight projection and refinement module $\psi(\cdot)$ maps the concatenated feature $[F^v_m;F^g_m]$ back to the original foundational feature space and predicts a residual update:
\begin{equation}
T_m = F^v_m + \psi([F^v_m; F^g_m]), \qquad m \in V,
\end{equation}
where $[\cdot;\cdot]$ denotes channel-wise concatenation.
This residual feature fusion injects cross-view geometric context into the semantic representation while keeping the teacher close to the original feature space.
We refine the teacher by optimizing the fusion module and LoRA adapters~\cite{hu2022lora} on the 2D backbone, while keeping the geometry model frozen.

\subsection{Discriminative Feature Learning}
\label{sec:feature_learning}

The fused teacher from \cref{sec: teacher_construction} combines semantic 2D features with multi-view geometric context, but feature fusion alone does not guarantee local feature separability.
We therefore encourage geometrically valid correspondences to rank above incompatible candidates in feature-similarity space.
For each query pixel $x\in\mathcal{Q}$, we sample a candidate set $\mathcal{C}(x)$ from other views using the available multi-view geometry.
Unreliable candidates are removed using reprojection validity and visibility/depth consistency.
Among the remaining candidates, binary relevance labels are assigned by 3D proximity: candidates with $\|X_x-X_y\|_2<\tau_{\mathrm{pos}}$ are positives, candidates with $\|X_x-X_y\|_2>\tau_{\mathrm{neg}}$ are negatives, and candidates in between are treated as ambiguous and removed from $\mathcal{C}(x)$.
Let $r_x(y)\in\{0,1\}$ denote the relevance label of the resulting candidate set and $s(x,y)=\cos(t_x,t_y)$ denote the teacher-feature similarity.
We optimize this ranking constraint using a differentiable average-precision objective~\cite{brown2020smoothap,mef}.
We approximate the rank of each candidate $y\in\mathcal{C}(x)$ as
\begin{equation}
R_x(y)
=
1+
\sum_{\substack{z\in\mathcal{C}(x)\\ z\neq y}}
\sigma\left(
\frac{s(x,z)-s(x,y)}{\tau}
\right),
\label{eq:soft_rank}
\end{equation}
where $\sigma(\cdot)$ is the sigmoid function and $\tau$ controls the softness of the rank approximation.
The positive rank is computed using only relevant candidates:
\begin{equation}
R_x^+(y)
=
1+
\sum_{\substack{z\in\mathcal{C}(x)\\ z\neq y}}
r_x(z)
\sigma\left(
\frac{s(x,z)-s(x,y)}{\tau}
\right).
\label{eq:positive_rank}
\end{equation}
We compute the ranking loss only over query pixels with at least one valid positive candidate,
$\mathcal{Q}^{+}=\{x\in\mathcal{Q}\mid\sum_{y\in\mathcal{C}(x)}r_x(y)>0\}$.
The discriminative loss is
\begin{equation}
\mathcal{L}_{\mathrm{disc}}
=
1-
\frac{1}{|\mathcal{Q}^{+}|}
\sum_{x\in\mathcal{Q}^{+}}
\frac{1}{Z_x}
\sum_{y\in\mathcal{C}(x)}
r_x(y)
\frac{R_x^+(y)}{R_x(y)},
\qquad
Z_x=
\sum_{y\in\mathcal{C}(x)} r_x(y)+\epsilon .
\label{eq:disc_loss}
\end{equation}
This loss penalizes cases where geometrically incompatible candidates are ranked above valid correspondences.
To allow geometric refinement while bounding deviation from the original 2D feature space, we use a semantic anchor to the frozen foundational feature $v_x$:
\begin{equation}
\mathcal{L}_{\mathrm{anchor}}
=
\frac{1}{|\mathcal{Q}|}
\sum_{x\in\mathcal{Q}}
\max\left(0, d_{\cos}(t_x,v_x)-\delta\right),
\label{eq:anchor_loss}
\end{equation}
where $d_{\cos}(a,b)=1-\cos(a,b)$ and $\delta$ allows limited deviation from the original representation.
The teacher refinement objective is
\begin{equation}
\mathcal{L}_{\mathrm{teacher}}
=
\mathcal{L}_{\mathrm{disc}}
+
\lambda_{\mathrm{anchor}}\mathcal{L}_{\mathrm{anchor}} .
\label{eq:teacher_loss}
\end{equation}

\subsection{Distilling Multi-view Priors into a Single-view Student}
\label{sec:distillation}

The refined teacher relies on multi-view inputs and geometry-based positive and negative candidates during training,
whereas our final goal is a practical encoder that operates on a single image.
After teacher refinement, we freeze the teacher and distill its representation into a single-view student.
For each selected view $m \in V$, the student independently processes the image and produces a feature map $S_m = f_\theta(I_m)$ in the same feature space as the teacher.
We supervise the student with feature-level distillation from the refined teacher while stopping gradients through the teacher:
\begin{equation}
\mathcal{L}_{\mathrm{distill}}
=
\frac{1}{|V|}
\sum_{m\in V}
\frac{1}{|\Omega_m|}
\sum_{x\in\Omega_m}
\frac{1}{D}
\left\|
\bar{S}_m(x)-\mathrm{sg}\big[\bar{T}_m(x)\big]
\right\|_1,
\label{eq:distill_loss}
\end{equation}
where $\bar{S}_m(x)$ and $\bar{T}_m(x)$ are the L2-normalized student and teacher features, $D$ is the feature dimension, $\Omega_m$ denotes the valid pixel set, and $\mathrm{sg}[\cdot]$ denotes stop-gradient.

To preserve the discriminative structure learned by the teacher, we also apply the same geometry-supervised ranking objective from Sec.~\ref{sec:feature_learning} to the student features as a weak regularizer.
The full student objective is
\begin{equation}
\mathcal{L}_{\mathrm{student}}
=
\mathcal{L}_{\mathrm{distill}}
+
\lambda_{\mathrm{disc}}\mathcal{L}_{\mathrm{disc}},
\label{eq:student_loss}
\end{equation}
where $\mathcal{L}_{\mathrm{disc}}$ is computed using the student features $S_m$, and $\lambda_{\mathrm{disc}}$ controls the strength of the discriminative regularization.
At inference time, only the student encoder is used, enabling standard single-image feature extraction with multi-view awareness distilled from the teacher.
\section{Experiments}
\label{sec:result}

We evaluate DDMS from three complementary perspectives: direct feature quality, dense prediction transfer, and explicit 3D lifting and aggregation.
Together, these experiments test whether our representation improves multi-view consistency and discriminability while preserving the semantic utility of pretrained visual features.
We compare against recent 3D-aware feature learning baselines and ablate the main components of our method.

\subsection{Experimental Setup}

\begin{table*}[t]
\centering
\setlength{\tabcolsep}{10pt}
\resizebox{\textwidth}{!}{
\begin{tabular}{@{}>{\centering\arraybackslash}m{0.1em} >{\centering\arraybackslash}m{9.5em}|G|MM|MM|MM|MM}
\toprule
\multicolumn{2}{c|}{}
& \multicolumn{1}{c|}{Separation}
& \multicolumn{4}{c|}{Geometric correspondence}
& \multicolumn{4}{c}{Semantic matching} \\
\cmidrule(lr){3-3} \cmidrule(lr){4-7} \cmidrule(l){8-11}
& Method
& ScanNet~\cite{dai2017scannet}
& \multicolumn{2}{c|}{ScanNet~\cite{dai2017scannet}}
& \multicolumn{2}{c|}{Navi-W~\cite{jampani2023navi}}
& \multicolumn{2}{c|}{DAVIS~\cite{davis2017}}
& \multicolumn{2}{c}{SPair-71k~\cite{min2019spair}} \\
\cmidrule(lr){3-3} \cmidrule(lr){4-5} \cmidrule(lr){6-7} \cmidrule(lr){8-9} \cmidrule(l){10-11}
&
& Margin$\uparrow$
& @10px$\uparrow$
& @20px$\uparrow$
& @0.05$\uparrow$
& @0.1$\uparrow$
& \multicolumn{2}{c|}{J\&F$\uparrow$}
& @0.05$\uparrow$
& @0.1$\uparrow$ \\
\midrule
\multirow{6}{*}{\rotatebox[origin=c]{90}{\scriptsize ViT-B}}
& DINOv2 {\scriptsize (baseline)}
& 0.4108
& 21.72
& 36.11
& 57.97
& 75.94
& \multicolumn{2}{c|}{67.82}
& 52.98
& 68.56 \\
& Fit3D {\scriptsize (ECCV'24)}
& 0.3494
& 22.89
& 39.50
& 31.26
& 53.81
& \multicolumn{2}{c|}{66.70}
& 30.25
& 44.86 \\
& MEF {\scriptsize (ICLR'25)}
& 0.4288
& \cellcolor{second}39.17
& \cellcolor{second}55.06
& \cellcolor{best}\textbf{67.71}
& \cellcolor{best}\textbf{80.87}
& \multicolumn{2}{c|}{66.82}
& \cellcolor{second}56.78
& \cellcolor{second}72.07 \\
& SnD {\scriptsize (ICLR'26)}
& \cellcolor{second}0.4520
& 28.65
& 45.67
& 59.53
& 76.86
& \multicolumn{2}{c|}{68.05} 
& 53.68
& 69.65 \\
& SnD {\scriptsize (+blending)}
& 0.3526
& 24.37
& 40.79
& 56.33
& 75.11
& \multicolumn{2}{c|}{\cellcolor{best}\textbf{69.11}}
& 52.25
& 68.14 \\
& Ours
& \cellcolor{best}\textbf{0.4632}
& \cellcolor{best}\textbf{66.01}
& \cellcolor{best}\textbf{78.48}
& \cellcolor{second}65.66
& \cellcolor{second}79.40
& \multicolumn{2}{c|}{\cellcolor{second}68.95}
& \cellcolor{best}\textbf{61.52}
& \cellcolor{best}\textbf{75.46} \\
\midrule
\multirow{6}{*}{\rotatebox[origin=c]{90}{\scriptsize ViT-S}}
& DINOv2 {\scriptsize (baseline)}
& 0.3967
& 23.00
& 38.16
& 53.72
& 73.03
& \multicolumn{2}{c|}{65.78}
& 51.11
& 67.16 \\
& Fit3D {\scriptsize (ECCV'24)}
& 0.3001
& 22.27
& 39.29
& 26.60
& 47.87
& \multicolumn{2}{c|}{66.01}
& 21.12
& 34.21 \\
& MEF {\scriptsize (ICLR'25)}
& \cellcolor{second}0.4170
& \cellcolor{second}38.54
& \cellcolor{second}54.30
& \cellcolor{best}\textbf{61.67}
& \cellcolor{best}\textbf{76.31}
& \multicolumn{2}{c|}{66.16}
& \cellcolor{second}51.64
& 67.32 \\
& SnD {\scriptsize (ICLR'26)}
& 0.3819
& 28.65
& 46.24
& 54.13
& 73.30
& \multicolumn{2}{c|}{67.10} 
& 51.13
& \cellcolor{second}67.83 \\
& SnD {\scriptsize (+blending)}
& 0.3147
& 23.26
& 39.39
& 50.22
& 70.42
& \multicolumn{2}{c|}{\cellcolor{best}\textbf{68.20}}
& 49.43
& 66.17 \\
& Ours
& \cellcolor{best}\textbf{0.4314}
& \cellcolor{best}\textbf{61.35}
& \cellcolor{best}\textbf{74.24}
& \cellcolor{second}59.57
& \cellcolor{second}74.55
& \multicolumn{2}{c|}{\cellcolor{second}67.68}
& \cellcolor{best}\textbf{54.09}
& \cellcolor{best}\textbf{68.70} \\
\bottomrule
\end{tabular}
}
\caption{%
Direct feature evaluation compared with recent 3D-aware feature finetuning methods~\cite{fit3d,mef,snd}.
Benchmarks cover ScanNet~\cite{dai2017scannet} feature separation and geometric correspondence, Navi-Wild~\cite{jampani2023navi} object correspondence, SPair-71k~\cite{min2019spair} semantic keypoint matching, and DAVIS~\cite{davis2017} mask propagation.
Best and second-best results are highlighted within each backbone group.
Our method achieves the best separation, outperforms existing methods by a large margin on ScanNet, and performs strongly in all cases.}

\label{tab:main_corr}
\end{table*}
\begin{figure*}[t]
\centering
\includegraphics[width=\linewidth]{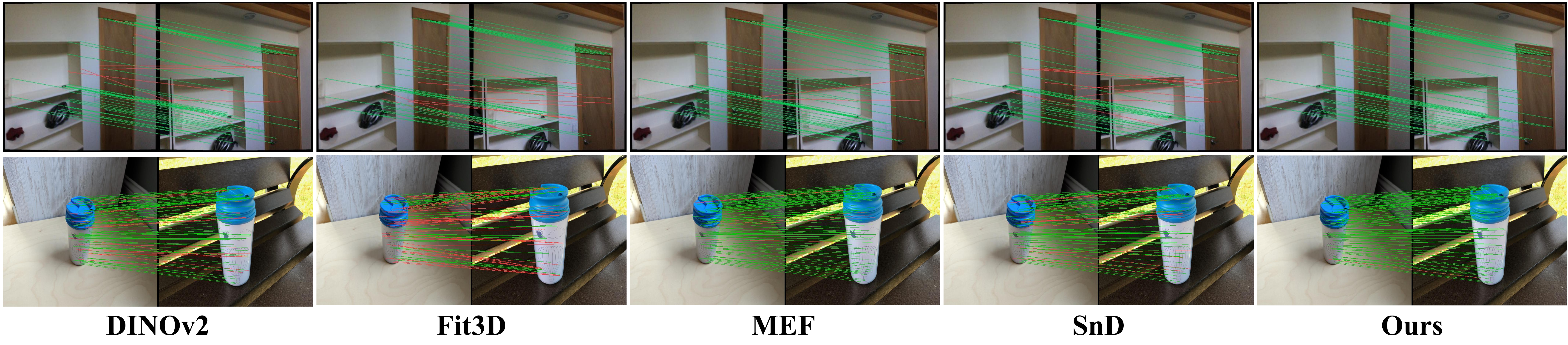}
\caption{%
Qualitative nearest-neighbor correspondence results on ScanNet~\cite{dai2017scannet} and Navi-Wild~\cite{jampani2023navi}.
Green and red lines denote correct and incorrect matches, respectively.
Our features produce cleaner correspondences with fewer mismatches in these examples.
}
\label{fig:qual_corr}
\end{figure*}
\paragraph{Benchmarks and evaluation.}
We first evaluate feature quality directly through feature separation, geometric and semantic correspondence, and temporal propagation on ScanNet~\cite{dai2017scannet}, Navi-Wild~\cite{jampani2023navi}, SPair-71k~\cite{min2019spair}, and DAVIS-2017~\cite{davis2017}.
Second, we evaluate downstream transferability with frozen-feature linear probes for semantic segmentation on ScanNet and ScanNet++~\cite{yeshwanth2023scannet++}, and depth estimation on ScanNet and NYUv2~\cite{silberman2012indoor_nyuv2}.
Third, we evaluate the features after explicit 3D lifting in two complementary settings. To isolate cross-view agreement, we perform point-based feature aggregation for ScanNet 3D point segmentation. We then lift image features onto pretrained 3D Gaussians~\cite{kerbl20233dgs,marrie2025ludvig} and render them from held-out views on the Mip-NeRF 360 scenes~\cite{barron2022mipnerf}.
Detailed evaluation protocols are provided in the supplementary appendix.

\paragraph{Baselines.}
For the main comparisons, we use DINOv2~\cite{oquab2023dinov2} as the common pretrained 2D backbone, since closely related 3D-aware feature finetuning and distillation methods provide DINOv2-based checkpoints.
We compare against frozen DINOv2, Fit3D~\cite{fit3d},
MEF~\cite{mef}, and SnD~\cite{snd}.
For SnD, we report both released checkpoints:
the checkpoint without SAM-mask blending, denoted SnD,
and the checkpoint with blending, denoted SnD (+blending).
This controlled setting keeps the initial feature space fixed and isolates how each method adapts the same pretrained representation for 3D-aware feature learning.

\begin{wrapfigure}{r}{0.58\textwidth}
    \centering
    \includegraphics[width=0.58\textwidth]{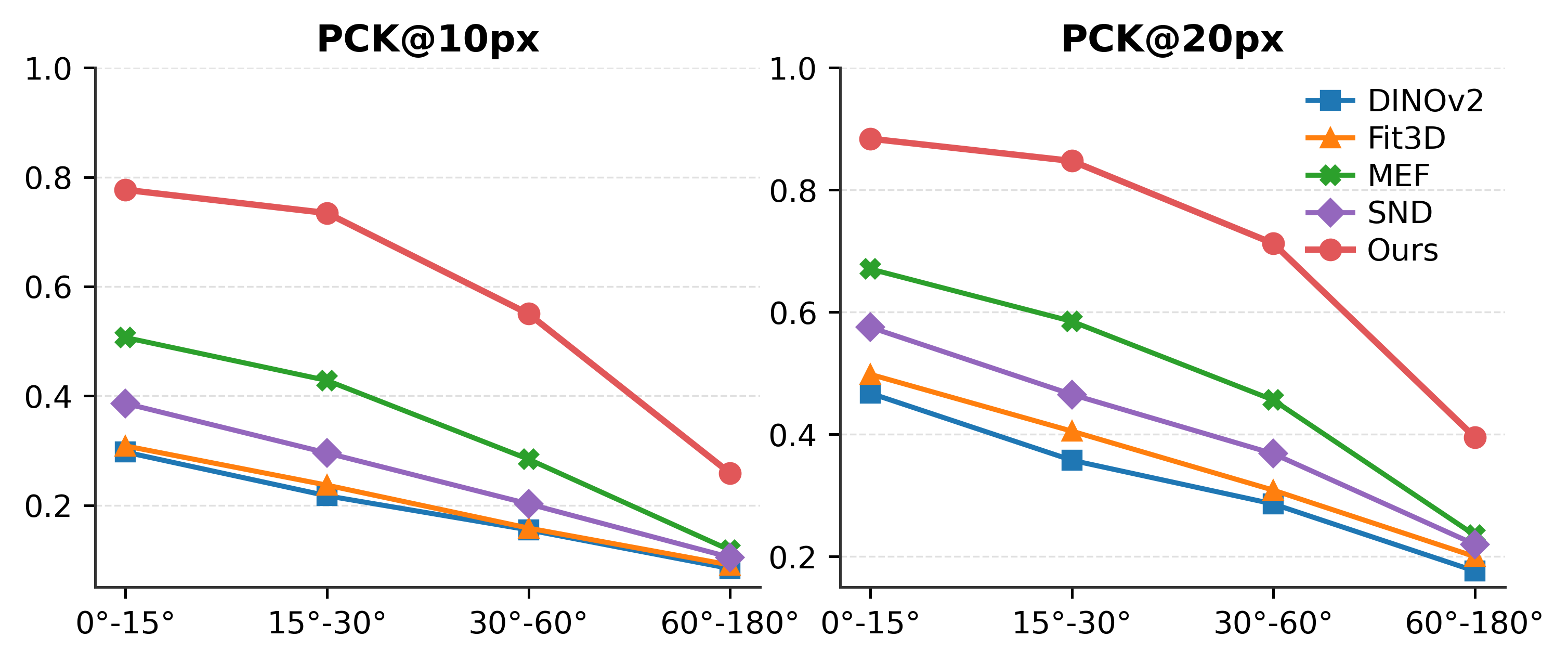}
    \caption{%
    Viewpoint-gap analysis for ScanNet~\cite{dai2017scannet} correspondence (ViT-B, PCK@10px and PCK@20px).
    Our method remains robust as viewpoint changes increase.
    }
    \label{fig:mv_corr_wrap}
\end{wrapfigure}

\subsection{Geometric and Semantic Feature Quality -- \Cref{tab:main_corr,fig:qual_corr,fig:mv_corr_wrap}}
\label{sec:feature_quality}

We first evaluate the learned representation in feature space, focusing on separation, geometric correspondence, semantic correspondence, and temporal propagation.

\paragraph{Feature separation.}
To measure feature distinctiveness, we compute the margin between the cosine similarity of each corresponding pair and the average similarity to non-corresponding points.
As shown in \Cref{tab:main_corr}, our method achieves the largest margin across both backbone variants, indicating enhanced separation.

\paragraph{Geometric correspondence.}
We first evaluate multi-view matching on ScanNet~\cite{dai2017scannet}, where image pairs
of indoor scenes are provided, and we measure the accuracy of the nearest-neighbor matches in feature space---the accuracy reflects geometric alignment.
As shown in \Cref{tab:main_corr}, our method achieves much higher
correspondence accuracy on ScanNet.
Qualitative results in \Cref{fig:qual_corr} also show more accurate matches with fewer errors.
The viewpoint-gap analysis in \cref{fig:mv_corr_wrap} further shows that our advantage persists as the viewpoint gap increases.

\paragraph{Semantic correspondence.}
We then test whether this improvement transfers beyond shared-scene geometry in \Cref{tab:main_corr}.
Navi-Wild~\cite{jampani2023navi} evaluates object-centric correspondence under appearance and context changes, while SPair-71k~\cite{min2019spair} measures semantic correspondence across object instances.
We also report DAVIS-2017~\cite{davis2017} as an auxiliary temporal propagation benchmark.
Across these settings, our method remains consistently strong, suggesting that the gains in geometric correspondence and feature separation do not come at the expense of semantic correspondence.

\begin{figure*}[b]
\centering
\includegraphics[width=\linewidth]{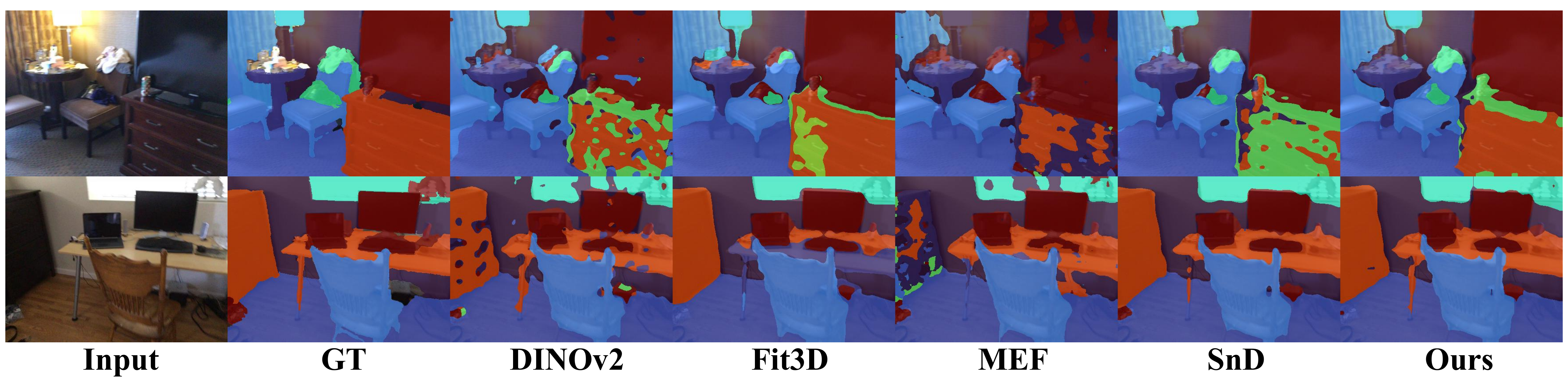}
\caption{%
Qualitative comparison of semantic segmentation using frozen features with linear probing.
}
\label{fig:qual_seg}
\end{figure*}

\begin{figure}[t]
    \centering
    \includegraphics[width=0.8\linewidth]{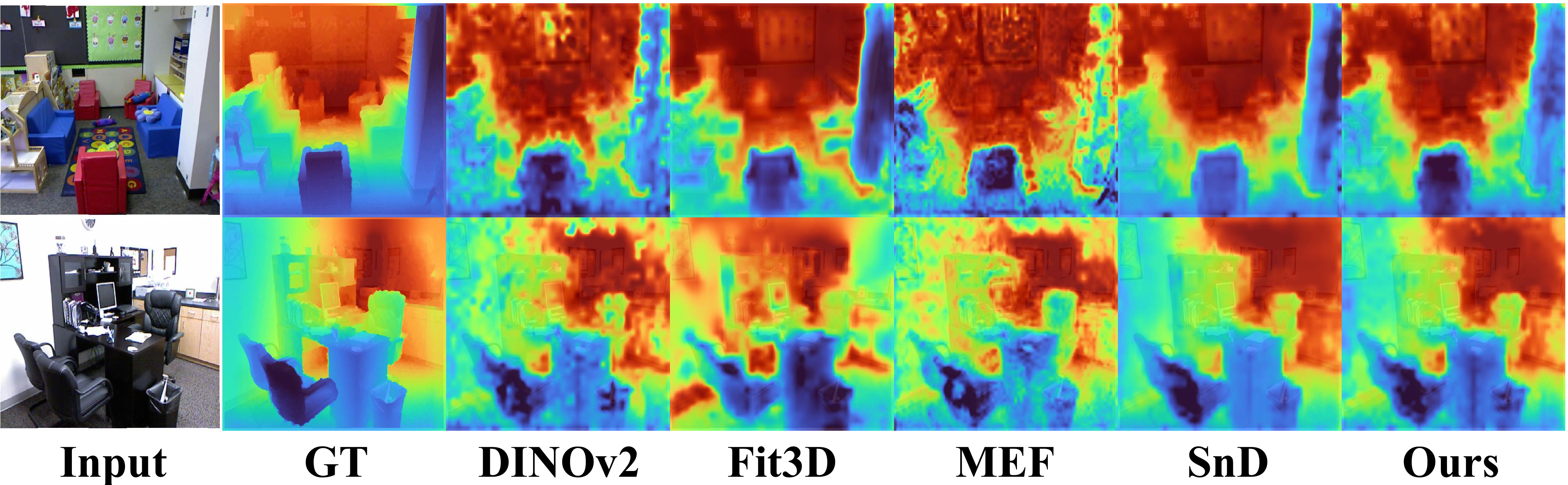}
    \caption{%
    Qualitative comparison of depth estimation using frozen features with linear probing.
    Our method provides coherent depth estimates.}
    \label{fig:qual_depth}
\end{figure}

\subsection{Dense Prediction Transfer via Linear Probing -- \Cref{tab:linear_probe,fig:qual_seg,fig:qual_depth}}

We next evaluate whether the learned features preserve downstream utility beyond correspondence-based tasks by performing linear probing for semantic segmentation and depth estimation.

\paragraph{Semantic segmentation.}
To probe semantic richness, we follow the standard protocol of fitting a linear transformation from our features to semantics.
As shown in \cref{fig:qual_seg}, our method produces clean segmentation maps, consistent with the quantitative results in \Cref{tab:linear_probe}.
Our method performs comparably to SnD (+blending) in semantic segmentation, while outperforming both SnD variants on ScanNet geometric correspondence in \Cref{tab:main_corr}.
DDMS therefore retains strong dense semantic transfer while improving
geometric correspondence.

\begin{table*}[t]
\setlength{\tabcolsep}{4pt}
\begin{center}
\resizebox{\linewidth}{!}{
\begin{tabular}{@{}>{\centering\arraybackslash}m{0.95em} lcccccc|cccccc}
\toprule
\multicolumn{2}{c}{}
& \multicolumn{6}{c|}{Semantic segmentation}
& \multicolumn{6}{c@{}}{Depth estimation} \\
\cmidrule(lr){3-8} \cmidrule(l){9-14}
&
& \multicolumn{3}{c}{ScanNet~\cite{dai2017scannet}}
& \multicolumn{3}{c|}{ScanNet++~\cite{yeshwanth2023scannet++}}
& \multicolumn{3}{c}{ScanNet~\cite{dai2017scannet}}
& \multicolumn{3}{c@{}}{NYUv2~\cite{silberman2012indoor_nyuv2}} \\
\cmidrule(lr){3-5} \cmidrule(lr){6-8}
\cmidrule(lr){9-11} \cmidrule(l){12-14}
& Method
& mIoU$\uparrow$ & mAcc$\uparrow$ & aAcc$\uparrow$
& mIoU$\uparrow$ & mAcc$\uparrow$ & aAcc$\uparrow$
& AbsRel$\downarrow$ & RMSE$\downarrow$ & $\delta_1\uparrow$
& AbsRel$\downarrow$ & RMSE$\downarrow$ & $\delta_1\uparrow$ \\
\midrule
\multirow{6}{*}{\rotatebox[origin=c]{90}{\scriptsize ViT-B}}
& DINOv2~\cite{oquab2023dinov2} & 53.69 & 66.29 & 77.63 & 31.69 & 41.41 & 81.47 & 0.1553 & 0.3718 & 78.12 & 0.1705 & 0.6566 & 74.90 \\
& Fit3D~\cite{fit3d} & 51.35 & 63.37 & 76.96 & 30.58 & 38.31 & 84.06 & 0.2045 & 0.4599 & 66.70 & 0.1975 & 0.7426 & 68.03 \\
& MEF~\cite{mef} & 52.49 & 65.17 & 76.93 & 29.40 & 38.62 & 81.43 & 0.1677 & 0.3937 & 75.06 & 0.1661 & 0.6222 & 75.98 \\
& SnD~\cite{snd} & 56.36 & 69.40 & 79.43 & 31.95 & 41.67 & 82.62 & \cellcolor{best}\textbf{0.1449} & \cellcolor{best}\textbf{0.3446} & \cellcolor{best}\textbf{80.89} & 0.1537 & 0.6018 & 78.56 \\
& SnD~\cite{snd} (+blending) & \cellcolor{best}\textbf{57.64} & \cellcolor{second}70.35 & \cellcolor{second}80.52 & \cellcolor{second}33.37 & \cellcolor{second}42.73 & \cellcolor{second}84.13 & 0.1454 & 0.3459 & \cellcolor{second}80.54 & \cellcolor{best}\textbf{0.1464} & \cellcolor{best}\textbf{0.5777} & \cellcolor{best}\textbf{80.70} \\
& Ours & \cellcolor{second}57.00 & \cellcolor{best}\textbf{70.51} & \cellcolor{best}\textbf{80.61} & \cellcolor{best}\textbf{33.53} & \cellcolor{best}\textbf{43.00} & \cellcolor{best}\textbf{84.34} & \cellcolor{second}0.1450 & \cellcolor{second}0.3453 & 80.48 & \cellcolor{second}0.1490 & \cellcolor{second}0.5835 & \cellcolor{second}79.78 \\
\midrule
\multirow{6}{*}{\rotatebox[origin=c]{90}{\scriptsize ViT-S}}
& DINOv2~\cite{oquab2023dinov2} & 50.18 & 62.81 & 75.06 & 27.80 & 37.06 & 79.52 & 0.1709 & 0.3950 & 74.70 & 0.1699 & 0.6465 & 75.26 \\
& Fit3D~\cite{fit3d} & 44.62 & 56.41 & 72.56 & 26.19 & 33.49 & 81.97 & 0.2171 & 0.4821 & 64.44 & 0.2120 & 0.7691 & 65.20 \\
& MEF~\cite{mef} & 47.22 & 59.70 & 73.29 & 24.53 & 32.96 & 78.69 & 0.1925 & 0.4319 & 69.72 & 0.1792 & 0.6541 & 73.39 \\
& SnD~\cite{snd} & 51.38 & 64.11 & 76.10 & 27.90 & 36.90 & 81.00 & \cellcolor{best}\textbf{0.1612} & \cellcolor{second}0.3780 & \cellcolor{best}\textbf{76.45} & \cellcolor{second}0.1536 & \cellcolor{second}0.5882 & \cellcolor{second}79.32 \\
& SnD~\cite{snd} (+blending) & \cellcolor{best}\textbf{52.61} & \cellcolor{best}\textbf{65.23} & \cellcolor{second}77.07 & \cellcolor{second}29.18 & \cellcolor{second}38.14 & \cellcolor{best}\textbf{82.40} & 0.1631 & \cellcolor{best}\textbf{0.3777} & 76.22 & \cellcolor{best}\textbf{0.1531} & \cellcolor{best}\textbf{0.5842} & \cellcolor{best}\textbf{79.46} \\
& Ours & \cellcolor{second}52.23 & \cellcolor{second}64.96 & \cellcolor{best}\textbf{77.12} & \cellcolor{best}\textbf{29.29} & \cellcolor{best}\textbf{38.53} & \cellcolor{second}82.17 & \cellcolor{second}0.1628 & 0.3786 & \cellcolor{second}76.25 & 0.1582 & 0.6036 & 78.13 \\
\bottomrule
\end{tabular}
}
\end{center}
\caption{%
Linear probing comparisons.
We evaluate semantic segmentation on ScanNet~\cite{dai2017scannet} and ScanNet++~\cite{yeshwanth2023scannet++}, and depth estimation on ScanNet~\cite{dai2017scannet} and NYUv2~\cite{silberman2012indoor_nyuv2}.
Our method is best or second-best on every segmentation metric and remains competitive with the SnD variants~\cite{snd} on depth, while outperforming both variants on ScanNet geometric correspondence in \Cref{tab:main_corr}.}
\label{tab:linear_probe}
\end{table*}

\paragraph{Depth estimation.}
To evaluate whether the learned representation retains single-view geometric cues, we train a linear depth probe on the frozen features.
As shown in \cref{fig:qual_depth}, our method produces coherent depth structure, consistent with the quantitative results in \Cref{tab:linear_probe}. 
DDMS remains competitive with SnD (+blending) on depth estimation while
outperforming both SnD variants on ScanNet correspondence in
\Cref{tab:main_corr}.
These results indicate that the learned representation retains useful
geometric cues while improving cross-view discriminability.

\begin{figure*}[t]
\centering
\includegraphics[width=\linewidth]{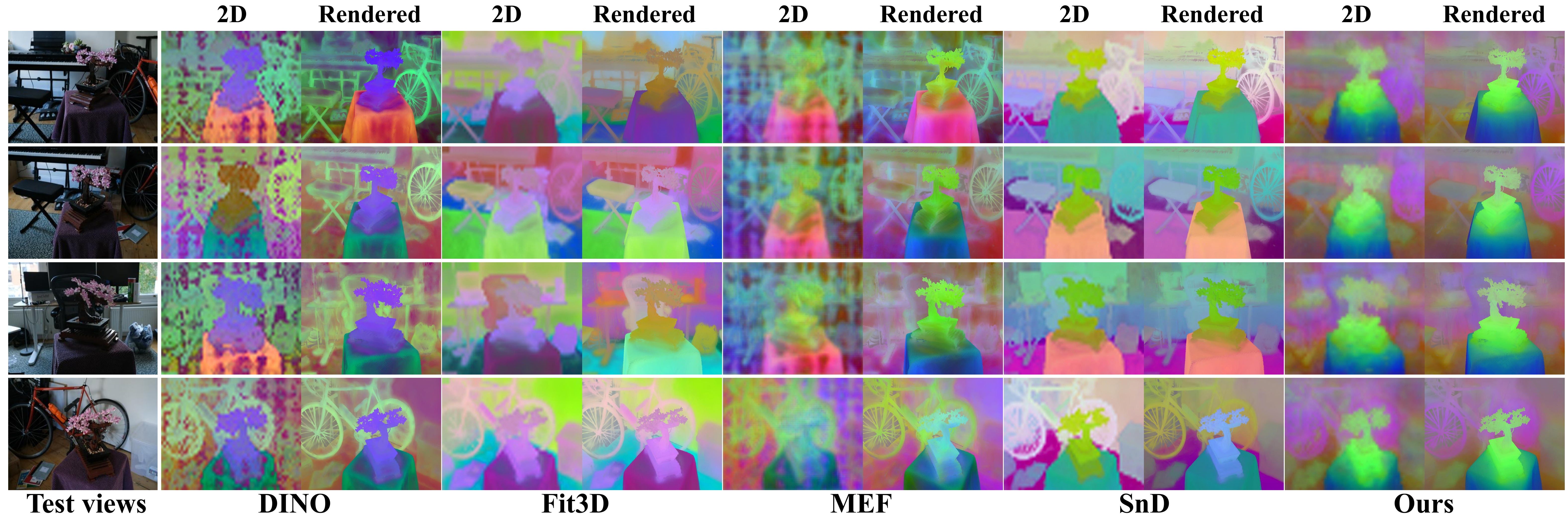}
\caption{%
Qualitative examples of feature uplifting and rendering on pretrained 3D Gaussians~\cite{barron2022mipnerf}.
PCA-projected 2D features are compared with rendered features using the same uplifting method~\cite{marrie2025ludvig}.
Our features remain more coherent across views and after rendering.
}
\label{fig:qual_feature_splat}
\end{figure*}

\begin{wrapfigure}{r}{0.5\textwidth}
    \centering
    \includegraphics[width=0.48\textwidth]{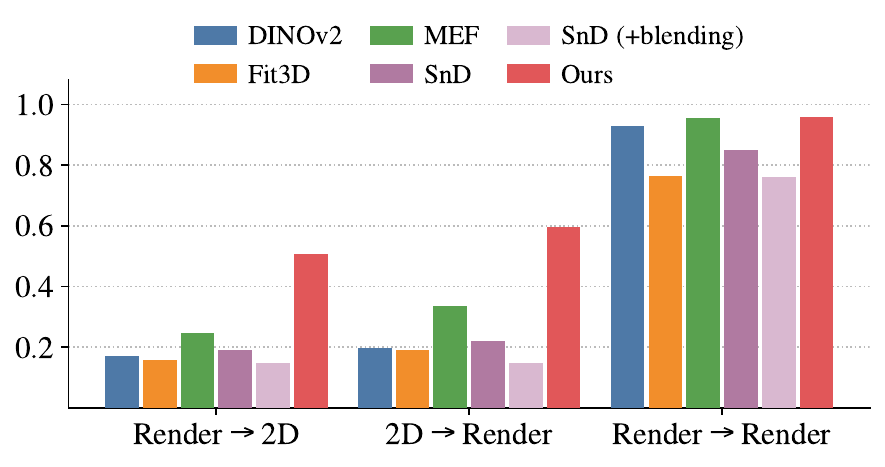}
    \caption{%
    Feature consistency between image features and rendered features, measured as nearest-neighbor matching accuracy (PCK@10px) when matching across them.
    Our method retains the highest accuracy, demonstrating consistency.
}
    \label{fig:feature_splat_graph}
\end{wrapfigure}

\subsection{Evaluating Features through 3D Lifting and Rendering -- \Cref{tab:point_lifting_seg,fig:qual_feature_splat,fig:feature_splat_graph}}
We further test whether the learned features remain consistent and discriminative after explicit 3D lifting and rendering.

\begin{table}[t]
\centering
\footnotesize
\setlength{\tabcolsep}{6pt}
\begin{tabular}{@{}lccccc}
\toprule
Method & mIoU$\uparrow$ & mAcc$\uparrow$ & aAcc$\uparrow$ & Per-view mIoU$\uparrow$ & Multi-view agree.$\uparrow$ \\
\midrule
DINOv2~\cite{oquab2023dinov2} & 62.30 & 74.30 & 82.69 & 54.49 & 77.34 \\
Fit3D~\cite{fit3d} & 59.54 & 71.04 & 81.45 & 54.21 & 80.73 \\
MEF~\cite{mef} & 61.54 & 73.73 & 82.24 & 54.07 & 77.71 \\
SnD~\cite{snd} & 62.79 & 74.57 & 82.99 & 56.51 & 80.28 \\
SnD~\cite{snd} (+blending) & \cellcolor{second}{63.26} & \cellcolor{best}\textbf{75.18} & \cellcolor{second}83.25 & \cellcolor{second}57.14 & \cellcolor{second}80.80 \\
Ours & \cellcolor{best}\textbf{63.32} & \cellcolor{second}74.85 & \cellcolor{best}\textbf{83.36} & \cellcolor{best}\textbf{57.79} & \cellcolor{best}\textbf{82.49} \\
\bottomrule
\end{tabular}
\caption{3D point segmentation on ScanNet~\cite{dai2017scannet} with a linear probe on features aggregated over views.
Per-view mIoU applies the same probe to the feature of each view without averaging, and multi-view agreement is the rate at which two views of the same point predict the same class.
Ours matches the strongest baseline after aggregation and is best per view, with the most consistent predictions across views.}
\label{tab:point_lifting_seg}
\end{table}

\paragraph{3D point segmentation.}
To directly look into 3D agreement,
we first evaluate semantic transfer after point-based 3D aggregation on ScanNet~\cite{dai2017scannet}.
We project image features directly onto 3D point clouds using posed RGB-D observations, average them across valid views, and train a linear classifier on the frozen aggregated features.
Since averaging over many views can hide disagreement between them, we also apply the same classifier to the feature of each individual view and measure how often two views of the same point agree.
As shown in \cref{tab:point_lifting_seg}, our method matches the strongest baseline after aggregation and performs best per view, while yielding the highest agreement across views. 
Importantly, this improved multi-view agreement is achieved while retaining the strong correspondence and dense-transfer performance observed in \cref{tab:main_corr,tab:linear_probe}.

\paragraph{3D feature splatting.}
We next evaluate the quality of features under a neural rendering setup.
Specifically, we evaluate feature splatting on pretrained 3D Gaussian scenes~\cite{kerbl20233dgs,barron2022mipnerf}.
Using a fixed uplifting method~\cite{marrie2025ludvig}, we assign image features to the already-trained 3D Gaussians of each scene
and render the resulting feature field from held-out views.
As shown in \cref{fig:qual_feature_splat},
several baselines show discrepancies between their original 2D features and rendered features after splatting, suggesting cross-view mismatch or smoothing during Gaussian feature accumulation.
In contrast, our rendered features remain better aligned with the corresponding 2D features and preserve coherent local structures.
We quantify this by measuring the accuracy of nearest-neighbor matches when matching between image features (2D) and rendered features in \cref{fig:feature_splat_graph}.
As expected, matching across rendered features shows high consistency for all methods, while these drop when matching rendered features to image features.
Ours retains the highest accuracy in all cases, demonstrating consistency between image features and uplifted features.

\subsection{Ablation study -- \Cref{tab:ablation}}
\begin{table}[t]
\footnotesize
\setlength{\tabcolsep}{8pt}

\begin{center}
\resizebox{0.58\linewidth}{!}{
\begin{tabular}{@{}lcccc}
\toprule
Variant & Sep. $\uparrow$ & Corr. $\uparrow$ & Sem. $\uparrow$ & Seg. $\uparrow$ \\
\midrule

DINOv2 (baseline)
& 0.4108 & 21.72 & 68.56 & 53.69 \\

Direct FT: $\mathcal{L}_{\mathrm{disc}}$
& \cellcolor{best}\textbf{0.5294}
& \cellcolor{best}\textbf{67.14}
& 57.36
& 31.44 \\

Direct FT: $\mathcal{L}_{\mathrm{disc}} + \mathcal{L}_{\mathrm{anchor}}$
& 0.4540
& 63.19
& 71.70
& 54.80 \\

w/o multi-view context
& 0.4535
& 62.23
& 70.51
& \cellcolor{second}56.71 \\

w/o teacher $\mathcal{L}_{\mathrm{anchor}}$
& 0.4619
& 65.14
& \cellcolor{second}73.45
& 54.93 \\

w/o student $\mathcal{L}_{\mathrm{disc}}$
& 0.4350
& 44.88
& 72.51
& 56.54 \\

Ours (full)
& \cellcolor{second}0.4632
& \cellcolor{second}66.01
& \cellcolor{best}\textbf{75.46}
& \cellcolor{best}\textbf{57.00} \\

\bottomrule
\end{tabular}
}
\end{center}

\caption{
Ablation study.
Sep./Corr. denote ScanNet feature separation and correspondence PCK@10px;
Sem./Seg. denote SPair-71k semantic correspondence PCK@0.1 and ScanNet segmentation mIoU.
}
\label{tab:ablation}
\end{table}
\label{sec:ablation}
To motivate our design choices, we report ablations in \cref{tab:ablation}, using ScanNet~\cite{dai2017scannet} feature separation and geometric correspondence, SPair-71k~\cite{min2019spair} semantic correspondence, and ScanNet semantic segmentation transfer.
Directly fine-tuning DINOv2 with $\mathcal{L}_{\mathrm{disc}}$ yields the strongest feature separation and geometric correspondence, but substantially degrades semantic correspondence and segmentation.
This reveals a clear trade-off: optimizing discriminability alone can over-specialize the representation toward geometric matching at the expense of its general semantic utility.
Adding $\mathcal{L}_{\mathrm{anchor}}$ to direct fine-tuning recovers much of the lost semantic performance while retaining strong geometric correspondence, but remains below the full DDMS model across all four metrics.
Starting from the full DDMS model, removing the teacher $\mathcal{L}_{\mathrm{anchor}}$, the student $\mathcal{L}_{\mathrm{disc}}$, or the multi-view context degrades performance across all four metrics.
In particular, removing the student discriminative objective causes a large drop in geometric correspondence, while removing the teacher anchor reduces both geometric and semantic performance.
Removing multi-view context also consistently reduces performance, confirming the benefit of cross-view geometric conditioning in the teacher.
Overall, the full model achieves the most favorable balance between geometric discriminability and semantic transfer, retaining strong separation and correspondence while achieving the best semantic correspondence and segmentation among the ablated variants.
Further ablations are provided in the supplementary appendix, including evaluations of off-the-shelf 3D-aware foundation features and different base models.

\section{Conclusion}
\label{sec:conclusion}

We presented DDMS, a 3D-aware feature distillation framework that transfers
multi-view geometric knowledge into a single-view visual encoder.
DDMS constructs a multi-view teacher through geometry-aware feature fusion,
discriminative ranking, and semantic anchoring, and distills the resulting
representation into a practical single-view student.
Across correspondence, dense prediction transfer, and explicit 3D lifting
evaluations, DDMS improves cross-view consistency and local discriminability
while preserving the semantic utility of the original foundation features.
Our results demonstrate that intermediate representations from multi-view
geometry models provide effective supervision for learning stronger
general-purpose 3D-aware visual features.
Despite requiring only a single image at inference time, DDMS relies on posed
multi-view RGB-D data and a geometry foundation model during training.
Moreover, because the distilled student does not perform explicit multi-view
reasoning at test time, it may remain limited when geometry is inherently
ambiguous from a single observation.
Exploring weaker geometric supervision and test-time multi-view extensions
provides promising directions for future work.
\clearpage
{
    \small
    \bibliographystyle{abbrvnat}
    \bibliography{ref}
}

\appendix

\clearpage

\section{Additional Analysis}
\label{app:additional_analysis}
\begin{figure*}[!b]
\centering
\includegraphics[width=\linewidth]{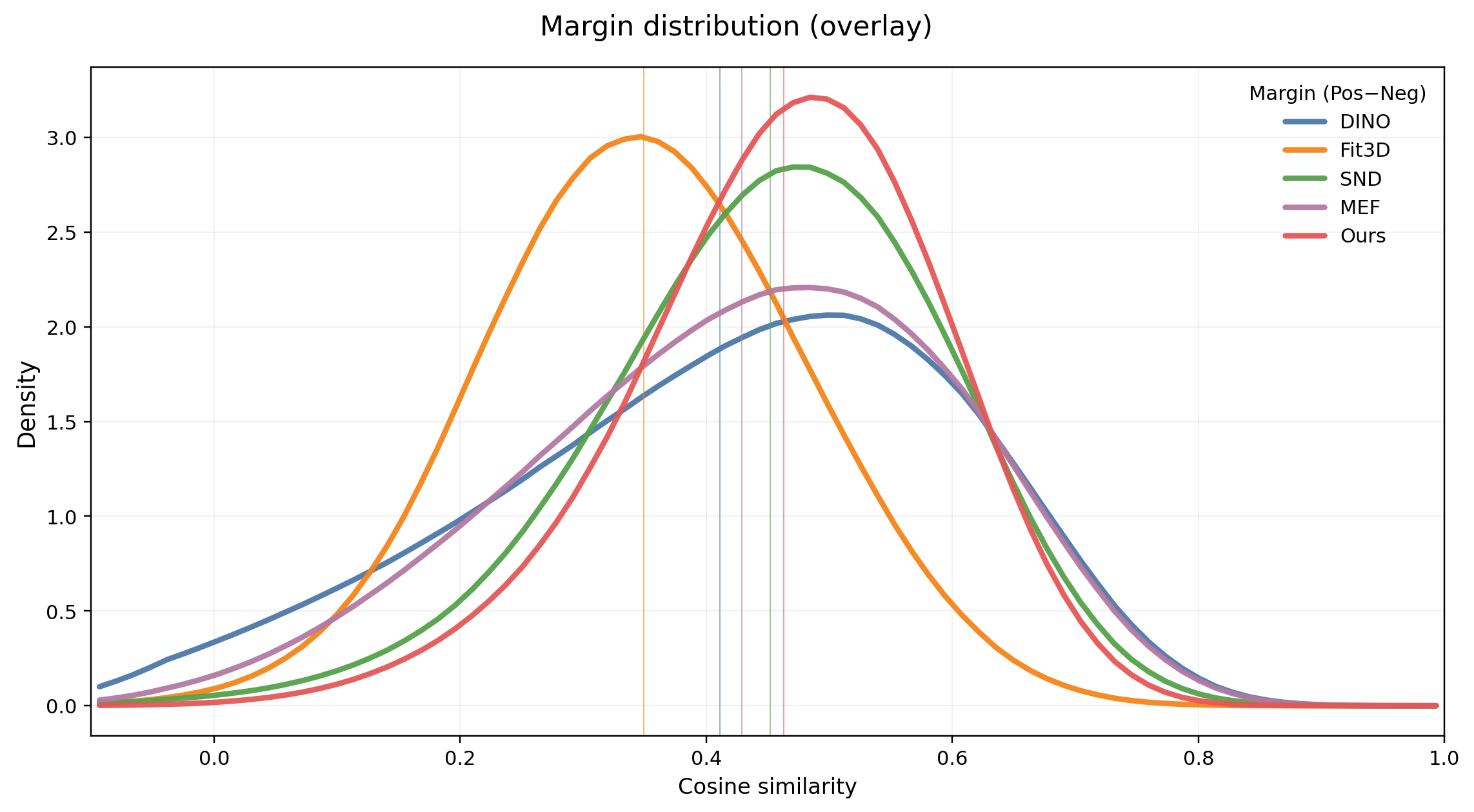}
\caption{
Distribution of per-query feature separation margins on the ScanNet~\cite{dai2017scannet} test pairs.
Vertical lines mark the mean of each method.
}
\label{fig:margin_dist}
\end{figure*}
We provide additional analyses to better understand the behavior and generality of the learned representation.
These analyses examine the distribution of feature separation, geometry feature selection, off-the-shelf 3D-aware foundation features, backbone generalization, and out-of-distribution transfer.

\subsection{Feature Separation Distribution}
\label{app:feature_separation_distribution}

We further analyze the feature separation results reported in \Cref{tab:main_corr}.
While the main table reports a single average margin over all ScanNet~\cite{dai2017scannet} test pairs from the SuperGlue~\cite{sarlin2020superglue} protocol, this aggregate score does not show how separation varies across individual queries.
We therefore analyze the per-query separation margins, using the same evaluation protocol, test pairs, query points, and negatives as in \Cref{tab:main_corr}.
\begin{wraptable}{r}{0.45\textwidth}
\begin{center}
\setlength{\tabcolsep}{6pt}
\renewcommand{\arraystretch}{1.05}
\resizebox{0.7\linewidth}{!}{%
\begin{tabular}{lcc}
\toprule
Method & Mean  & Std. \\
\midrule
DINOv2~\cite{oquab2023dinov2} & 0.4108 & 0.1935 \\
Fit3D~\cite{fit3d}          & 0.3494 & 0.1276 \\
SnD~\cite{snd}              & \cellcolor{second}0.4520 & 0.1388 \\
SnD~\cite{snd} (+blending)  & 0.3526 & \cellcolor{second}0.1233 \\
MEF~\cite{mef}              & 0.4288 & 0.1709 \\
Ours                        & \cellcolor{best}\textbf{0.4632} & \cellcolor{best}\textbf{0.1226} \\
\bottomrule
\end{tabular}%
}
\end{center}
\caption{
Mean and standard deviation of the per-query feature separation margins on the ScanNet test pairs.
}
\label{tab:margin_analysis}
\end{wraptable}
Each sample is the positive-negative margin of one valid query point, pooled over all 1,500 test pairs; the mean therefore equals the separation reported in \Cref{tab:main_corr}.
\Cref{tab:margin_analysis} reports the mean and standard deviation of these per-query margins, while \cref{fig:margin_dist} visualizes the full distribution.

Fit3D and SnD (+blending) produce relatively compact distributions, but their margins are centered at lower values.
DINOv2 and MEF achieve higher mean margins than these two, but show a much wider spread, with a heavier tail of low-margin queries.
SnD without blending reaches a higher mean margin than DINOv2 and MEF with a moderate spread.
In contrast, our method shifts the distribution toward larger margins and achieves the highest mean margin while maintaining the lowest standard deviation.
This indicates that the proposed distillation improves feature separation consistently across queries, rather than through a small subset of highly distinctive points, supporting the intended balance between cross-view consistency and local discriminability.

\subsection{Choice of geometry feature layer}
\label{sec:layer_choice}
\begin{table}
\begin{center}
\setlength{\tabcolsep}{6pt}
\renewcommand{\arraystretch}{1.08}
\resizebox{0.35\linewidth}{!}{%
\begin{tabular}{llcc}
\toprule
Block & Stage & Sep. $\uparrow$ & Corr. $\uparrow$ \\
\midrule
13 & Early  & 0.3302 & 27.94 \\
19 & Middle & \cellcolor{best}\textbf{0.3682} & \cellcolor{best}\textbf{37.03} \\
39 & Final  & 0.0860 & 13.54 \\
\bottomrule
\end{tabular}%
}
\end{center}
\caption{
DA3 feature layer analysis.
Frozen DA3-Giant features from early, middle, and final encoder blocks, evaluated directly on ScanNet feature separation and correspondence (PCK@10px).
}
\label{tab:da3_layer_ablation}
\end{table}
Our teacher uses intermediate features from Depth-Anything-3 (DA3)~\cite{lin2025da3}, a frozen geometry foundation model, as geometric context for refining pretrained 2D foundation features.
We therefore compare features from early, middle, and final stages of DA3's cross-view reasoning process, evaluating the frozen features directly with the separation and correspondence protocol of \Cref{tab:main_corr}.
Specifically, block 13 corresponds to the first layer where cross-view attention is applied, block 19 represents an intermediate cross-view feature, and block 39 is the final layer of the DA3 encoder.
As shown in \Cref{tab:da3_layer_ablation}, the middle-stage feature gives the best separation and correspondence performance.
This suggests that teacher construction benefits from features that have undergone sufficient cross-view interaction, but are not yet overly specialized to the final geometry prediction head.
Early features provide weaker geometric context, likely because they have only begun cross-view reasoning, while final features perform poorly as dense descriptors due to their specialization for the output geometry task.

\subsection{Evaluation of Off-the-Shelf 3D-Aware Foundation Features}
\label{app:gfm_features}

We evaluate whether existing 3D-aware foundation features can directly serve as general-purpose dense visual representations.
For VGGT~\cite{wang2025vggt} and Depth-Anything-3 (DA3)~\cite{lin2025da3}, we use their commonly adopted large-scale variants and evaluate the frozen multi-view features directly.
DUNE~\cite{sariyildiz2025dune} is a single-view universal encoder distilled from heterogeneous 2D and 3D teachers~\cite{oquab2023dinov2,wang2024dust3r,baradel2024multihmr}, and we evaluate its frozen dense representation.
This comparison evaluates whether strong off-the-shelf 3D-aware representations are already sufficient as dense visual features.
We report ScanNet feature separation (Sep.) and correspondence PCK@10px (Corr.), SPair-71k semantic correspondence PCK@0.1 (Sem.), and ScanNet semantic segmentation mIoU (Seg.).
\begin{table}[t]
\begin{center}
\setlength{\tabcolsep}{3pt}
\renewcommand{\arraystretch}{1.08}
\resizebox{0.80\linewidth}{!}{%
\begin{tabular}{llllccccc}
\toprule
Model & Scale & Params & Dim. & Mode & Sep. $\uparrow$ & Corr. $\uparrow$ & Sem. $\uparrow$ & Seg. $\uparrow$ \\
\midrule
VGGT~\cite{wang2025vggt}
& ViT-L + Agg. & 1.26B & 2048 & Multi
& 0.3096 & 36.98 & \cellcolor{second}65.89 & \cellcolor{second}53.26 \\
DA3~\cite{lin2025da3}
& ViT-G & 1.14B & 1536 & Multi
& \cellcolor{second}0.3682 & \cellcolor{second}37.03 & 48.74 & 39.90 \\
\midrule
DUNE~\cite{sariyildiz2025dune}
& ViT-B & 86M & 768 & Single
& 0.1846 & 33.10 & 59.52 & 47.76 \\
\midrule
Ours
& ViT-B & 92M & 768 & Single
& \cellcolor{best}\textbf{0.4632}
& \cellcolor{best}\textbf{66.01}
& \cellcolor{best}\textbf{75.46}
& \cellcolor{best}\textbf{57.00} \\
\bottomrule
\end{tabular}%
}
\end{center}
\caption{
Evaluation of off-the-shelf 3D-aware foundation features.
All features are evaluated frozen; Mode indicates multi-view (Multi) or single-view (Single) inference.
}
\label{tab:gfm_features}
\end{table}
As shown in \Cref{tab:gfm_features}, off-the-shelf 3D-aware features are not automatically strong general-purpose dense features.
Although VGGT and DA3 are much larger geometry foundation models with multi-view inference, DDMS achieves the strongest performance on all four metrics with a lighter single-view encoder.
Compared with DUNE, which has a comparable parameter count and single-view inference mode, DDMS also shows stronger feature performance across separation, correspondence, semantic matching, and segmentation.
These results suggest that geometry-aware features are most effective when used as context for refining and distilling pretrained 2D foundation features, rather than being used directly as the final representation.

\subsection{Downstream Camera Pose Estimation}
\label{app:pose_estimation}

We further evaluate whether the improved feature correspondence transfers to downstream camera pose estimation on ScanNet~\cite{dai2017scannet}, 7-Scenes~\cite{7scene}, and ETH-3D~\cite{schops2017eth3d}.
For each dataset, evaluation pairs are fixed across methods, and query poses are estimated using the same mutual nearest-neighbor matching and PnP-RANSAC pipeline.
Ground-truth depth is used to construct the reference 3D points so that the comparison isolates descriptor quality.
We report pose recall R@10 and R@25, the percentage of queries localized within $10$\,cm/$10^\circ$ and $25$\,cm/$10^\circ$ of the ground-truth pose.
As shown in \Cref{tab:pose_estimation}, DDMS achieves the highest recall on all three datasets, improving substantially over both the original DINOv2 features and the 3D-aware baselines, showing that the learned representation transfers from feature-level correspondence to a downstream geometric task.
\begin{table}[t]
\centering
\small
\setlength{\tabcolsep}{5pt}
\renewcommand{\arraystretch}{1.05}

\begin{tabular}{lccc}
\toprule
Feature & ScanNet & 7Scenes & ETH3D \\
\midrule
DINOv2~\cite{oquab2023dinov2} & 15.1 / 26.9 & 76.1 / 90.3 & 16.7 / 36.5 \\
Fit3D~\cite{fit3d} & 19.0 / 38.0 & 77.3 / 92.9 & 16.0 / 38.8 \\
MEF~\cite{mef} & \cellcolor{second}30.7 / 48.1 & \cellcolor{second}88.3 / 98.4 & \cellcolor{second}26.2 / 54.0 \\
SnD~\cite{snd} & 20.6 / 36.9 & 81.1 / 92.3 & 15.2 / 36.9 \\
SnD~\cite{snd} (+blending) & 18.8 / 34.1 & 77.0 / 93.0 & 15.2 / 38.8 \\
Ours & \cellcolor{best}\textbf{65.9 / 85.4} & \cellcolor{best}\textbf{95.4 / 100.0} & \cellcolor{best}\textbf{49.0 / 80.6} \\
\bottomrule
\end{tabular}

\caption{
Downstream camera pose estimation with PnP-RANSAC.
Each entry reports pose recall R@10 / R@25 (\%), within $10$\,cm/$10^\circ$ and $25$\,cm/$10^\circ$.
}
\label{tab:pose_estimation}
\end{table}

\subsection{Generalization Across Backbones}
\label{app:backbone_generalization}

The main paper focuses on DINOv2~\cite{oquab2023dinov2} ViT-B and ViT-S for fair comparison with competitive baselines that provide public checkpoints in these settings.
We further evaluate DDMS with additional foundation backbones: DINOv2-L and DINOv2-B with registers~\cite{darcet2023vision} (DINOv2-L-reg, DINOv2-B-reg), and DINOv3-B~\cite{simeoni2025dinov3}.
These experiments test whether the proposed 3D-aware distillation strategy generalizes across backbone scales and pretraining variants.
As in \Cref{tab:gfm_features}, Sep. and Corr. denote ScanNet feature separation and correspondence PCK@10px, and Sem. and Seg. denote SPair-71k PCK@0.1 and ScanNet segmentation mIoU.
\begin{table}[t]

\begin{center}

\setlength{\tabcolsep}{5pt}
\renewcommand{\arraystretch}{1.08}

\resizebox{0.6\linewidth}{!}{%

\begin{tabular}{c l cccc}

\toprule
Backbone & Model & Sep. $\uparrow$ & Corr. $\uparrow$ & Sem. $\uparrow$ & Seg. $\uparrow$ \\
\midrule

\multirow{2}{*}{\scriptsize\centering ViT-L}
& DINOv2-L-reg &
0.4315 &
23.68 &
67.83 &
55.84 \\

& \hspace{1.5em}+ Ours &
\cellcolor{best}\textbf{0.5413} &
\cellcolor{best}\textbf{61.39} &
\cellcolor{best}\textbf{77.75} &
\cellcolor{best}\textbf{57.91} \\

\midrule

\multirow{2}{*}{\scriptsize\centering ViT-B}
& DINOv2-B-reg &
0.4167 &
26.61 &
68.21 &
54.67 \\

& \hspace{1.5em}+ Ours &
\cellcolor{best}\textbf{0.5127} &
\cellcolor{best}\textbf{60.74} &
\cellcolor{best}\textbf{75.72} &
\cellcolor{best}\textbf{55.72} \\

\midrule

\multirow{2}{*}{\scriptsize\centering ViT-B}
& DINOv3-B &
0.3251 &
30.26 &
69.93 &
\cellcolor{best}\textbf{57.10} \\

& \hspace{1.5em}+ Ours &
\cellcolor{best}\textbf{0.4455} &
\cellcolor{best}\textbf{64.76} &
\cellcolor{best}\textbf{73.85} &
56.92 \\

\bottomrule

\end{tabular}%

}

\end{center}

\caption{
Generalization across foundation backbones.
Each backbone is compared with DDMS trained on top of it (+ Ours), using the metrics of \Cref{tab:gfm_features}.
}

\label{tab:backbone_generalization}

\end{table}

As shown in \Cref{tab:backbone_generalization}, DDMS consistently improves feature separation, geometric correspondence, and semantic correspondence across all tested backbones. The gains are especially pronounced for geometric correspondence, indicating that the proposed distillation strategy is not tied to a specific DINOv2-B/S configuration. For semantic segmentation, DDMS yields clear improvements on the DINOv2 variants and remains comparable on DINOv3, where the change is marginal. Overall, these results suggest that DDMS enhances 3D-aware feature quality across different foundation backbones while largely preserving dense semantic transfer.

\subsection{Out-of-Distribution Dense Prediction Transfer}
\label{app:ood_transfer}
\begin{figure*}[!b]
\centering
\includegraphics[width=0.9\linewidth]{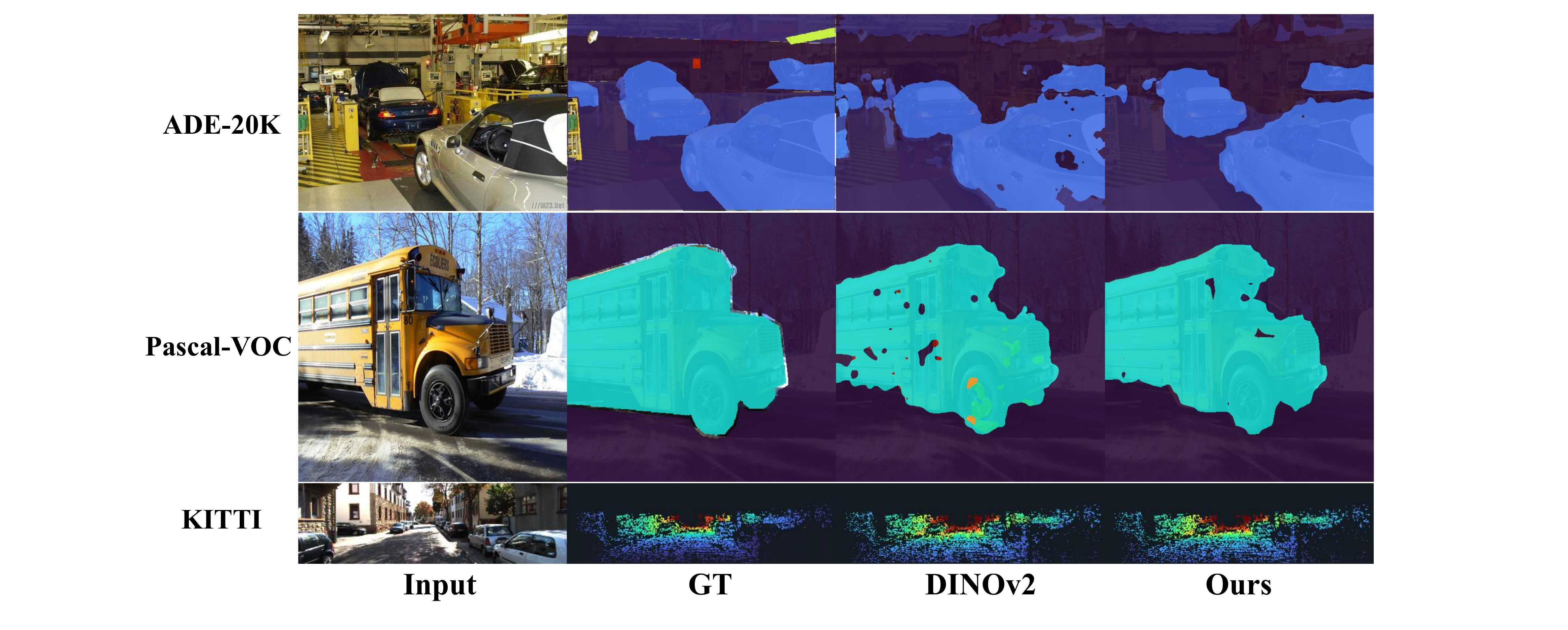}
\caption{
Qualitative out-of-distribution transfer with frozen-feature linear probes: semantic segmentation on ADE20K~\cite{zhou2017scene_ade20k} and Pascal VOC~\cite{everingham2010pascal}, and depth estimation on KITTI~\cite{geiger2013vision_kitti}.
}
\label{fig:ood_transfer}
\end{figure*}

\cref{fig:ood_transfer} shows qualitative examples from additional dense prediction datasets beyond the indoor RGB-D domains used in the main transfer evaluation.
While the main dense prediction results focus primarily on indoor scene understanding, we further evaluate frozen-feature linear probes on ADE20K~\cite{zhou2017scene_ade20k} and Pascal VOC~\cite{everingham2010pascal} for semantic segmentation, and KITTI~\cite{geiger2013vision_kitti} for outdoor depth estimation, using the probe protocol of the main paper (KITTI images are processed at 1008$\times$294 to retain sufficient vertical patch resolution).
These datasets cover broader scene categories, object-centric images, and outdoor driving scenes, testing whether DDMS preserves general-purpose transferability beyond the indoor domain.
The quantitative results in \Cref{tab:ood_transfer} show that DDMS continues to perform well in these out-of-distribution settings.
\begin{table}
\centering
\small
\setlength{\tabcolsep}{4pt}
\renewcommand{\arraystretch}{1.08}
\resizebox{0.9\linewidth}{!}{%
\begin{tabular}{lccccccccc}
\toprule
& \multicolumn{3}{c}{ADE-20K~\cite{zhou2017scene_ade20k}}
& \multicolumn{3}{c}{Pascal-VOC~\cite{everingham2010pascal}}
& \multicolumn{3}{c}{KITTI~\cite{geiger2013vision_kitti}} \\
\cmidrule(lr){2-4} \cmidrule(lr){5-7} \cmidrule(lr){8-10}
Model
& mIoU $\uparrow$ & mAcc $\uparrow$ & aAcc. $\uparrow$
& mIoU $\uparrow$ & mAcc $\uparrow$ & aAcc. $\uparrow$
& RMSE $\downarrow$ & AbsRel $\downarrow$ & $\delta_1$ $\uparrow$ \\
\midrule
DINOv2 & 44.29 & 55.45 & 79.67 & 81.30 & 87.52 & 95.85 & 4.0027 & 0.1228 & 85.90 \\
Ours   & \cellcolor{best}\textbf{47.14} & \cellcolor{best}\textbf{57.81} & \cellcolor{best}\textbf{82.15} & \cellcolor{best}\textbf{84.97} & \cellcolor{best}\textbf{90.16} & \cellcolor{best}\textbf{96.82} & \cellcolor{best}\textbf{3.7735} & \cellcolor{best}\textbf{0.1207} & \cellcolor{best}\textbf{86.17} \\
\bottomrule
\end{tabular}%
}
\caption{
Out-of-distribution dense prediction transfer.
We evaluate frozen-feature linear probes on ADE-20K~\cite{zhou2017scene_ade20k} and Pascal-VOC~\cite{everingham2010pascal} for semantic segmentation, and on KITTI~\cite{geiger2013vision_kitti} for depth estimation.
}
\label{tab:ood_transfer}
\end{table}

\section{Additional Qualitative Results}
\label{app:additional_qual}

We provide additional qualitative examples from the evaluation settings used in the main paper.
These examples complement the quantitative results and illustrate the behavior of the learned representation across correspondence, dense prediction, and 3D aggregation tasks.

\paragraph{Geometric correspondence.}
\begin{figure*}
\centering
\includegraphics[width=0.95\linewidth]{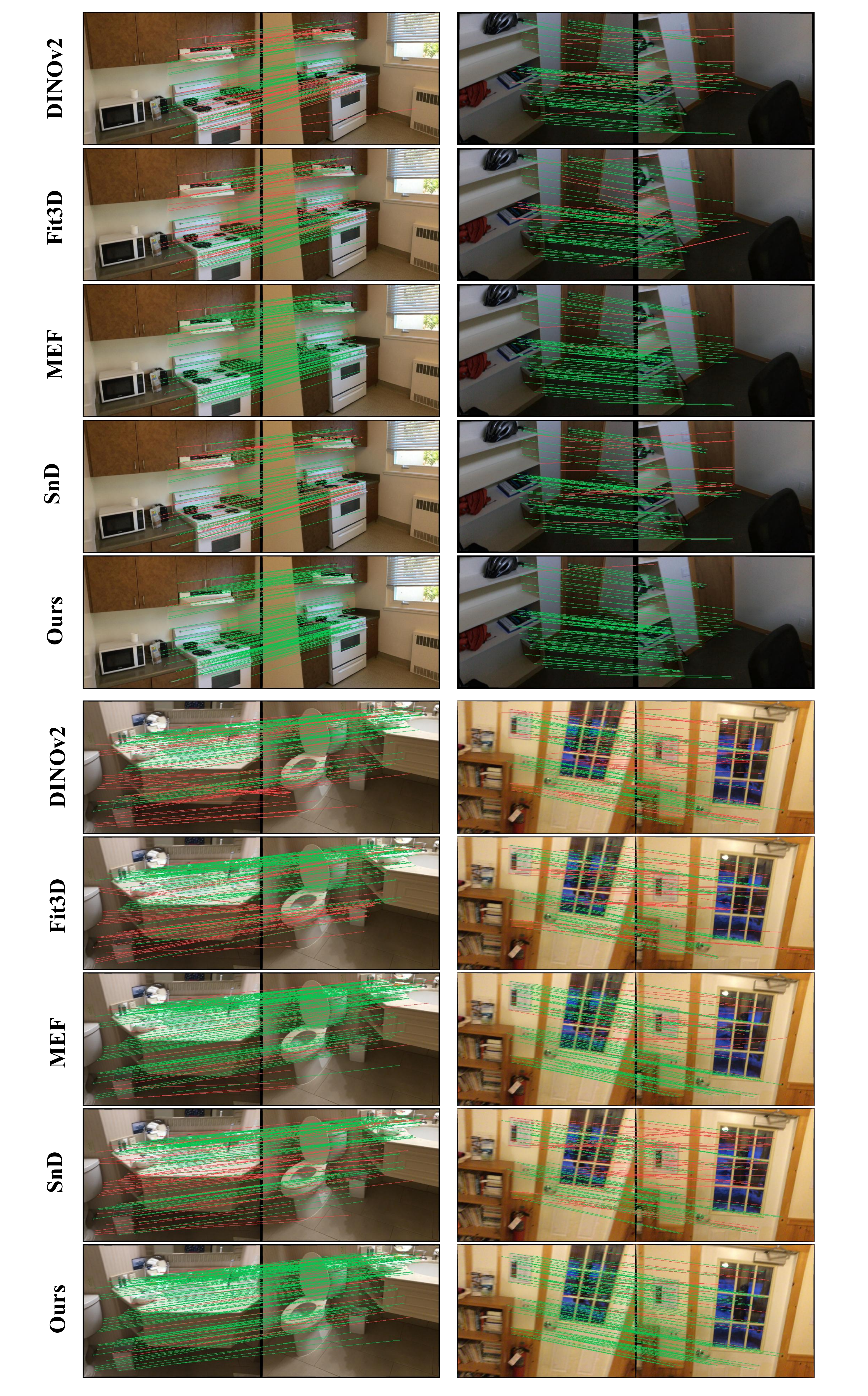}
\caption{
Additional nearest-neighbor correspondence results on ScanNet~\cite{dai2017scannet}, compared with all baselines.
Green and red lines indicate correct and incorrect matches.
}
\label{fig:sup_qual_corr_scannet}
\end{figure*}
\cref{fig:sup_qual_corr_scannet} provides additional nearest-neighbor matching examples on ScanNet~\cite{dai2017scannet}.
These examples evaluate feature matching between different views of the same indoor scene, where correct matches require geometric consistency and local discriminability.

\paragraph{In-the-wild correspondence.}
\begin{figure*}[!t]
\centering
\includegraphics[width=\linewidth]{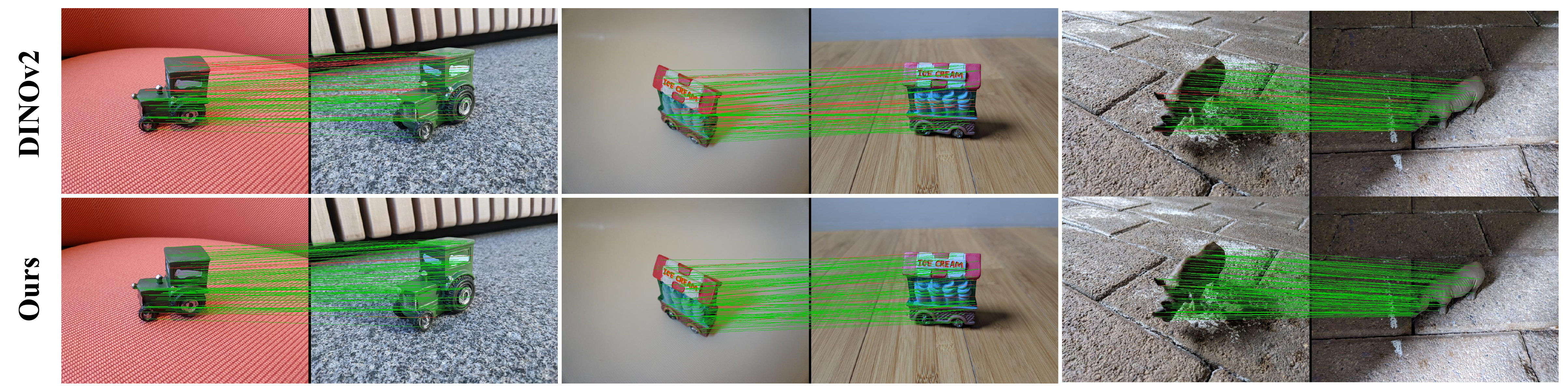}
\caption{
Additional correspondence results on Navi-Wild~\cite{jampani2023navi}, where the same object appears under different backgrounds and viewpoints.
DDMS is compared with DINOv2.
}
\label{fig:sup_qual_corr_navi}
\end{figure*}
\cref{fig:sup_qual_corr_navi} provides additional nearest-neighbor matching examples on Navi-Wild~\cite{jampani2023navi}.
Unlike ScanNet, these pairs contain the same object under different backgrounds, viewpoints, and visual contexts, testing whether the learned features remain robust beyond shared-scene geometry.

\begin{figure*}[t]
\centering
\includegraphics[width=\linewidth]{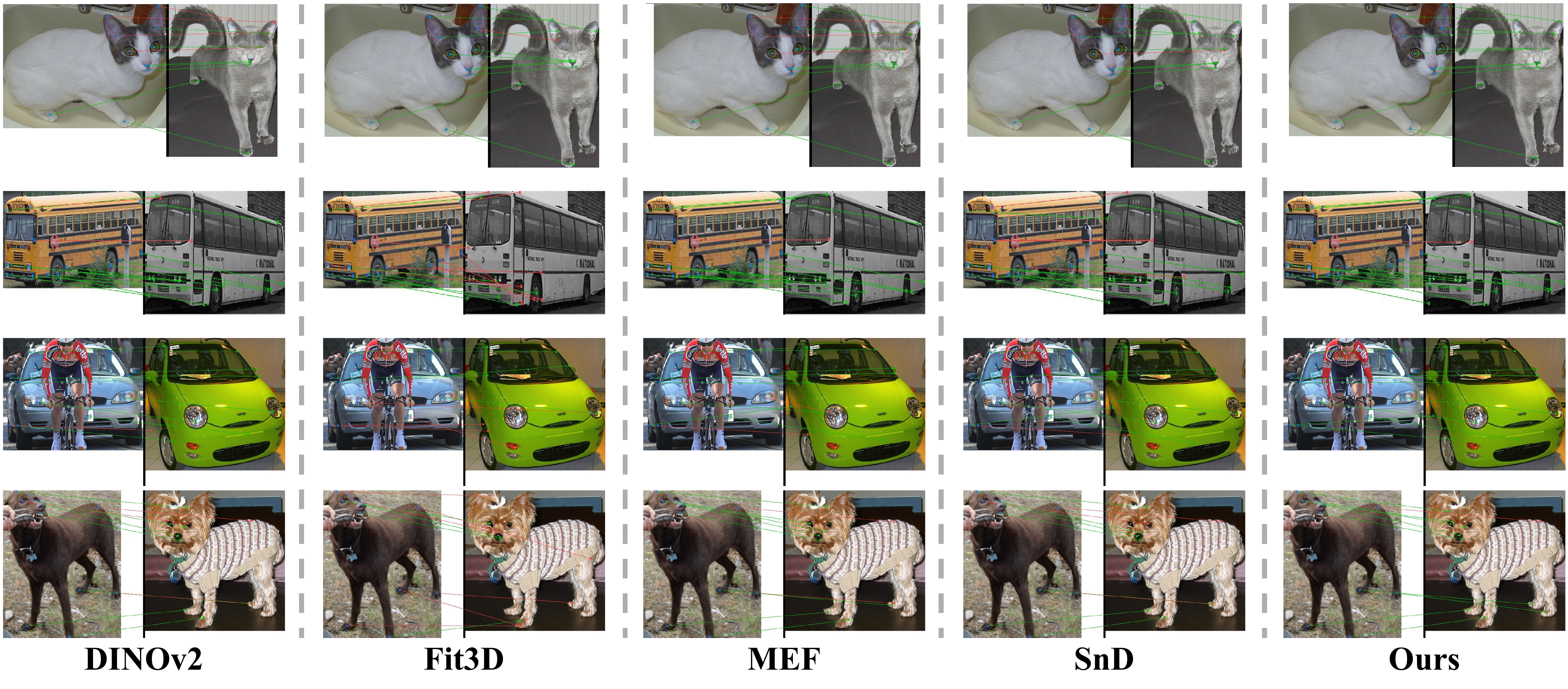}
\caption{
Additional semantic keypoint matching results on SPair-71k~\cite{min2019spair}, where keypoints are matched across different instances of the same category.
}
\label{fig:sup_sem_corr_spair}
\end{figure*}
\paragraph{Semantic keypoint matching.}
\cref{fig:sup_sem_corr_spair} provides additional semantic keypoint matching examples on SPair-71k~\cite{min2019spair}.
This benchmark matches keypoints across different object instances of the same category, testing whether the learned features preserve semantic correspondence ability beyond shared-scene geometry.

\begin{figure*}[t]
\centering
\includegraphics[width=\linewidth]{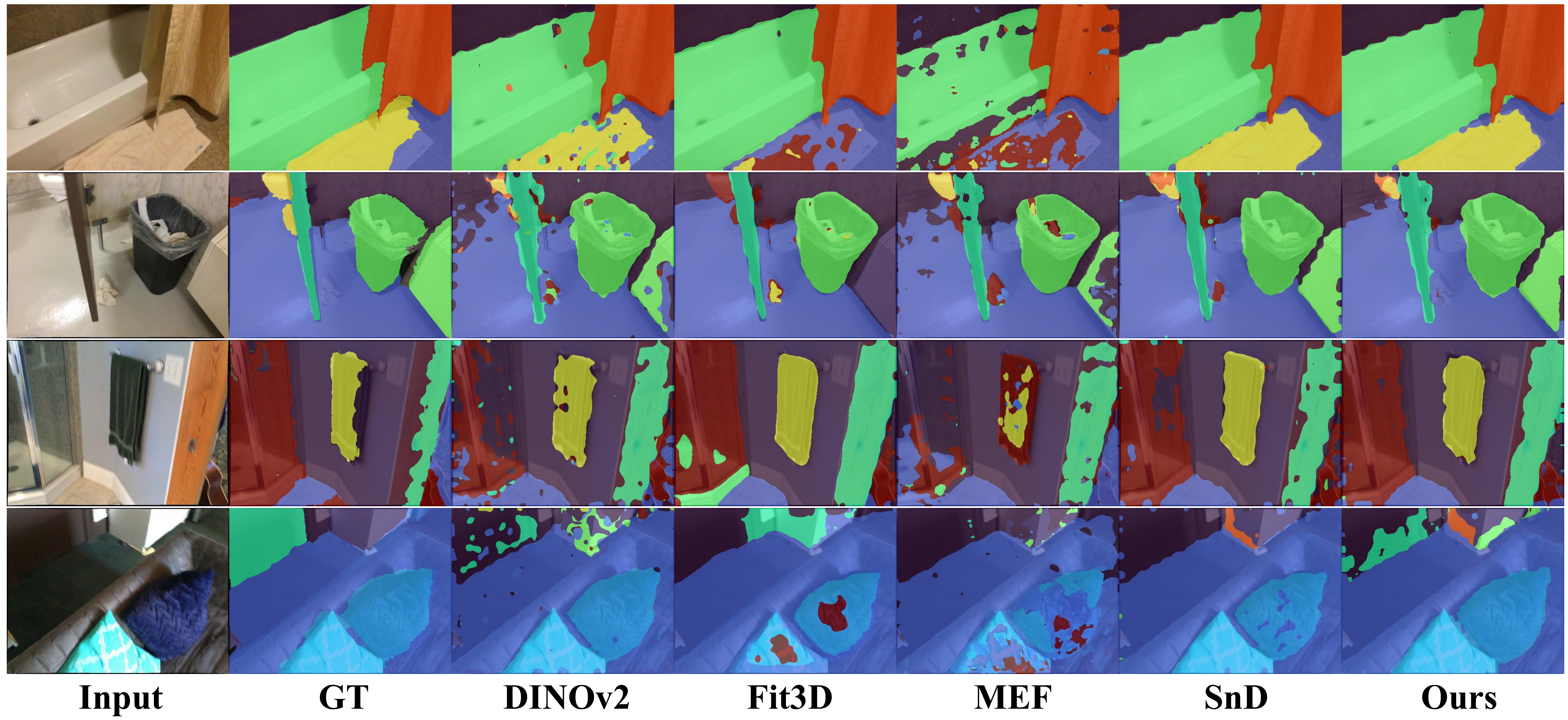}
\caption{
Additional semantic segmentation results from frozen-feature linear probing.
All methods use the same linear probe.
}
\label{fig:sup_qual_seg}
\end{figure*}
\paragraph{Semantic segmentation.}
\cref{fig:sup_qual_seg} provides additional semantic segmentation examples from frozen-feature linear probing.
These examples evaluate whether the learned representation preserves dense semantic structure for downstream transfer beyond correspondence-based tasks.

\begin{figure*}[t]
\centering
\includegraphics[width=0.8\linewidth]{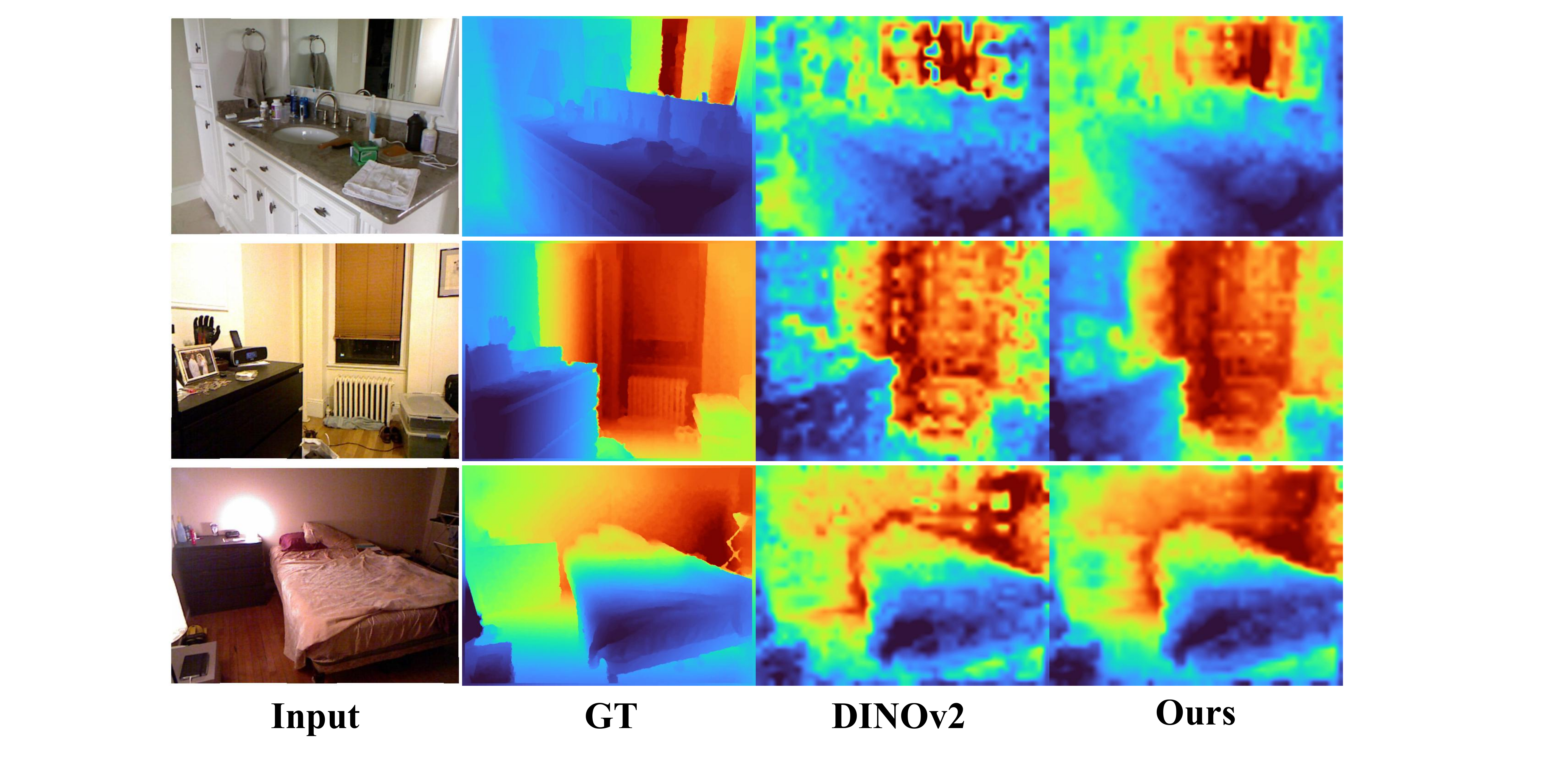}
\caption{
Additional depth estimation results from frozen-feature linear probing, comparing DDMS with DINOv2.
}
\label{fig:sup_qual_depth}
\end{figure*}

\paragraph{Depth estimation.}
\cref{fig:sup_qual_depth} provides additional depth estimation examples from frozen-feature linear probing.
We compare DDMS with the DINOv2 baseline to examine whether the learned representation preserves dense geometric cues under single-view inference.
The examples show that DDMS better preserves scene layout and object-level depth structure.

\begin{figure*}[t]
\centering
\includegraphics[width=0.95\linewidth]{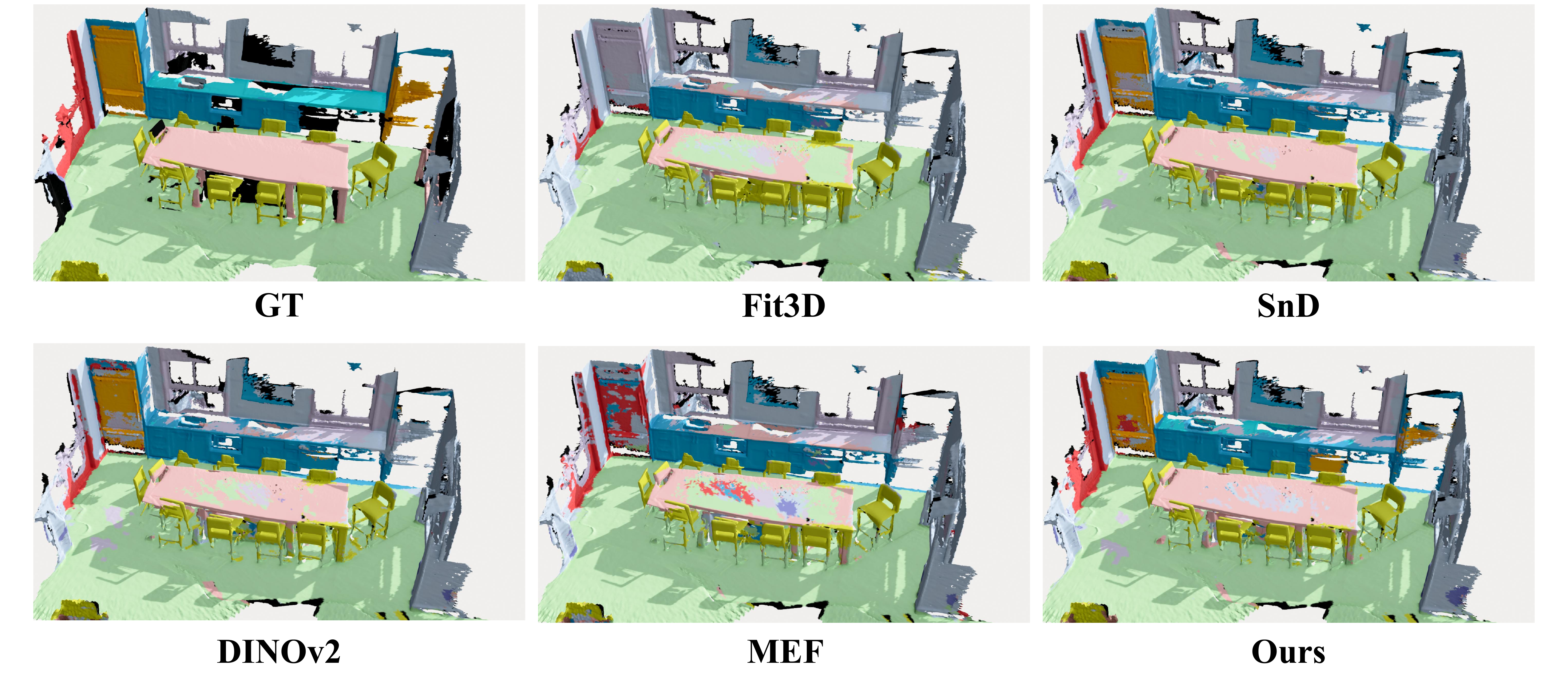}
\caption{
Additional 3D point segmentation results on ScanNet~\cite{dai2017scannet}.
Features are aggregated from posed RGB-D views onto the point cloud and classified with a linear probe.
}
\label{fig:sup_3d_point_seg}
\end{figure*}

\paragraph{3D point segmentation.}
\cref{fig:sup_3d_point_seg} provides additional qualitative results for point-based 3D feature aggregation.
After image features are lifted to 3D point clouds and evaluated with a linear probe, the results show whether the representation retains semantic structure after cross-view accumulation.

\begin{figure*}[t]
\centering
\includegraphics[width=\linewidth]{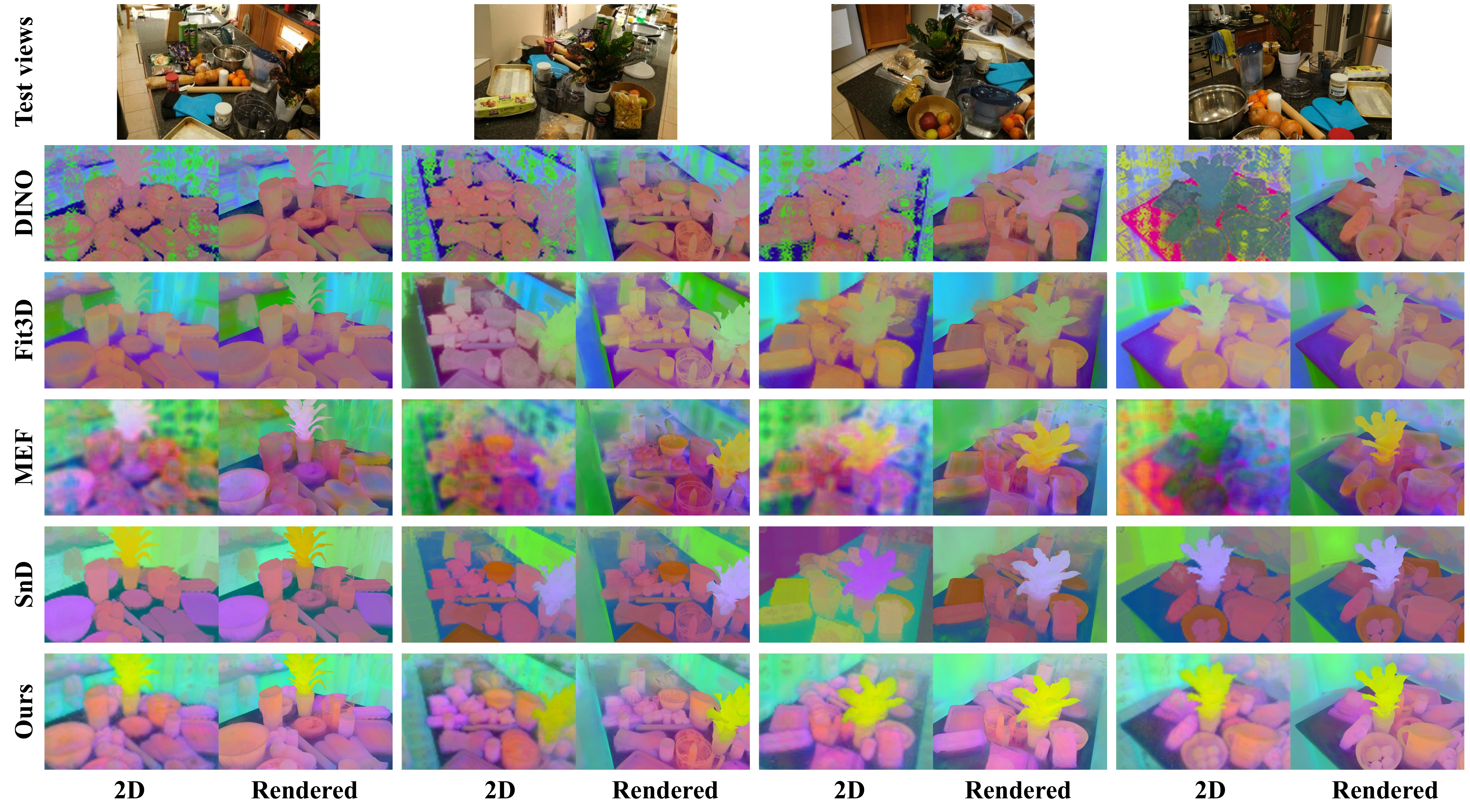}
\caption{
Additional feature splatting results on pretrained 3D Gaussian scenes.
For each method, the original 2D features are shown next to the features rendered from held-out views after uplifting.
}
\label{fig:sup_3d_feat_splat}
\end{figure*}
\paragraph{Feature splatting.}
\cref{fig:sup_3d_feat_splat} provides additional qualitative examples of feature splatting on pretrained 3D Gaussian scenes.
Image features are lifted to the Gaussians using the same fixed uplifting procedure as in the main paper and rendered from held-out novel views, testing whether the feature representation remains coherent after 3D aggregation and rendering.

\section{Implementation Details}
\label{sec:supp_training_impl}

\paragraph{Training setup.}
We train DDMS on posed multi-view RGB-D sequences from ScanNet++~\cite{yeshwanth2023scannet++}, using only training scenes and excluding evaluation scenes at the scene level.
Camera poses and depth maps are used only for constructing geometric supervision and candidate labels.
The final distilled student operates on a single image and does not require camera poses, depth maps, multi-view inputs, or the geometry model at inference time.

\paragraph{Teacher construction.}
We construct the multi-view teacher by fusing features from a pretrained 2D backbone with features from a frozen Depth-Anything-3 (DA3) model~\cite{lin2025da3}.
Unless otherwise specified, we use the feature from the 19th DA3-Giant encoder block, following the layer analysis in \Cref{tab:da3_layer_ablation}.
DA3 remains frozen throughout training and provides multi-view geometric context through its intermediate features.
For all teacher inputs, images are resized following the DA3 preprocessing, with the longer side set to 504 pixels while preserving the aspect ratio.
The 2D and geometry features are concatenated, projected with a 1 $\times$ 1 convolution, refined with a
3 $\times$ 3 convolution, and added back to the original 2D feature as a residual update.
This lightweight residual refinement head keeps the teacher representation in the original backbone feature space while injecting multi-view geometric context.
We train the teacher refinement stage for 20k steps using AdamW~\cite{loshchilov2019adamw} with a learning rate of $1\times10^{-4}$, weight decay of $1\times10^{-4}$, and gradient clipping with a maximum norm of 1.0.
Each iteration uses two multi-view samples, each with $N=4$ input views jointly processed by DA3.
For the ranking objective in \Cref{sec:method}, we sample $M=2$ views from the same scene and construct dense query-candidate matches.
The trainable parameters are the fusion module and LoRA~\cite{hu2022lora} adapters on the 2D backbone.
For each query pixel, positives and negatives are assigned from a pool of 1,024 candidate pixels.
Candidates within $0.05$m are treated as positives, candidates farther than $0.15$m as negatives, and candidates between the thresholds are ignored.
We further filter candidates using reprojection validity and visibility/depth consistency.
For the ranking loss, we use temperature $0.01$.
For the semantic anchor, we set $\lambda_{\mathrm{anchor}}=0.2$ and cosine-distance tolerance $\delta=0.2$.

\paragraph{Student distillation.}
After teacher refinement, we distill the refined multi-view teacher into a single-view student.
The student is initialized from the pretrained 2D foundation backbone and adapted with LoRA~\cite{hu2022lora} of rank 32.
The student also includes a lightweight $3{\times}3$ convolutional refinement head applied on top of the single-view backbone features.
For each training sample, the frozen teacher processes $N=4$ views jointly to produce refined multi-view teacher features.
We select $M=3$ views from this set, feed them independently to the student, and supervise the student with the objective in \Cref{eq:student_loss}, using $\lambda_{\mathrm{disc}}=0.2$.
For the distillation loss, $\Omega_m$ denotes all valid patch locations in the feature map after image resizing; geometric validity filtering is used only for the ranking objective.
One of the $M$ views serves as the reference view: feature distillation is applied to every view, and the discriminative loss to each pair formed by the reference view and one of the other views.
For the student ranking loss, we use a temperature of 0.02, positive/negative 3D-distance thresholds of 0.05\,m/0.10\,m, and a 15-pixel exclusion radius around the ground-truth match for negatives, with 1,024 additional random candidate pixels sampled from the paired view.
We tune the student for 20k steps using the same AdamW settings as teacher refinement with a cosine learning-rate schedule.
At inference time, the student operates on a single image without requiring multi-view inputs, camera poses, depth maps, or the DA3 geometry model.

\section{Evaluation Protocols}
All methods are evaluated with their publicly released weights and the same input resolution within each task (a 504-pixel longer side unless stated otherwise), using their own dense features without concatenating them with DINOv2 features.
SnD~\cite{snd} is reported with both of its released checkpoints:
the checkpoint without SAM-mask blending is denoted SnD,
and the checkpoint with blending is denoted SnD (+blending).

\subsection{Geometric and Semantic Feature Quality}

\paragraph{Feature separation.}
We evaluate feature separation on ScanNet~\cite{dai2017scannet} using the test image pairs from the SuperGlue~\cite{sarlin2020superglue} protocol.
For each pair, we use the first image as the reference view and sample query pixels on a regular grid.
Queries with invalid depth, invalid reprojection, or failed visibility/depth-consistency checks are discarded, and the same valid query set is used for all methods.
For each valid query, we reproject the 3D point into the paired view using ground-truth depth and camera poses to obtain the positive correspondence.
We then sample 50 random negative pixels in the paired view, excluding pixels within a 30-pixel radius of the positive location to avoid ambiguous nearby matches.
The same query, positive, and negative locations are used for all methods.
For each query, the separation margin is defined as the query-positive cosine similarity minus the average cosine similarity to the 50 negatives.
We report the margin averaged over all valid queries.

\paragraph{Geometric correspondence.}
We evaluate geometric correspondence on ScanNet~\cite{dai2017scannet} and Navi-Wild~\cite{jampani2023navi}.
For ScanNet, we use the test image pairs from the SuperGlue~\cite{sarlin2020superglue} protocol.
Building on the protocol of Probe3D~\cite{el2024probing}, we sample the dense features of each method on a common pixel grid at one quarter of the original image resolution ($324{\times}242$), and match every grid point with valid depth in the first view to its nearest neighbor in the second view.
Unlike Probe3D, we do not apply a ratio test or keep only the top-ranked matches, so all methods are scored on the same set of queries.
We report PCK@10px and PCK@20px on this grid, with ground-truth correspondences obtained from depth and camera poses; \cref{fig:mv_corr_wrap} reports the same metric binned by the relative rotation between the two views.

For Navi-Wild, we evaluate correspondence under in-the-wild object-centric conditions, where the target object appears across different contexts, viewpoints, and visual conditions.
We use image pairs from the Navi test set with relative viewpoint changes below $120^\circ$ and sample query points only inside the provided object mask.
We further discard queries whose ground-truth point is not visible in the target view, i.e., whose reprojection falls outside the target object mask or disagrees with the target depth by more than 5\%.
Following the Probe3D protocol~\cite{el2024probing}, correspondences are obtained by nearest-neighbor search in feature space and evaluated using PCK@0.05 and PCK@0.10 with normalized image coordinates.
This measures whether predicted correspondences fall within the specified normalized distance from the ground-truth match, testing robustness beyond shared-scene geometric matching.

\paragraph{Semantic correspondence.}
We evaluate semantic correspondence on SPair-71k~\cite{min2019spair} and DAVIS-2017~\cite{davis2017}.
For SPair-71k, we use the official test set and restrict evaluation to image pairs with view-difference level 0.
Given a source keypoint, we find its nearest neighbor in the target image in feature space and compare the predicted location with the annotated target keypoint.
We report PCK at the normalized thresholds used in the main paper, measuring the percentage of predicted correspondences that fall within the specified distance from the ground-truth keypoint.
This evaluates semantic matching across different object instances of the same category.

For DAVIS-2017, we evaluate temporal correspondence through video object mask propagation on the validation split.
Given the ground-truth mask in the first frame, labels are propagated to target frames using frozen backbone features without task-specific training.
Labels are transferred from memory-frame patches to target-frame patches based on feature similarity, and the predicted masks are evaluated with the standard DAVIS metrics: region similarity $\mathcal{J}$, boundary accuracy $\mathcal{F}$, and their average $\mathcal{J}\&\mathcal{F}$.
We report the average over two memory settings: first-frame-only propagation and propagation using the first frame plus the most recent past frame.

\subsection{Dense Prediction Transfer}
\label{sec:supp_protocol_dense_transfer}

\paragraph{Semantic segmentation.}
We evaluate semantic segmentation with linear probing on frozen dense features.
For each method, the backbone is fixed and only a one-layer spatial linear classifier is trained on top of the dense feature map.
The classifier predicts semantic logits at the feature-grid resolution, which are bilinearly upsampled to the ground-truth label resolution for loss computation and evaluation.
We train the probe with cross-entropy loss on valid labeled pixels, ignoring unlabeled or void regions.
We evaluate on ScanNet~\cite{dai2017scannet} and ScanNet++~\cite{yeshwanth2023scannet++}, using the official train/validation scene splits of both datasets.
All methods use the same splits, probe architecture, optimizer settings, and 50k-iteration training schedule.
We report mean intersection-over-union (mIoU), mean class accuracy (mAcc), and overall pixel accuracy (aAcc).

\paragraph{Depth estimation.}
We evaluate depth estimation with linear probing on frozen dense features, following the setup used in our Fit3D and SnD comparisons.
For each method, the backbone is fixed and only a one-layer depth classifier is trained.
Following common bin-classification depth estimation protocols~\cite{bhat2021adabins}, we concatenate the global \texttt{[CLS]} token to each spatial patch token, predict a distribution over discretized depth bins, and bilinearly upsample the logits before aligning them with the ground-truth depth map.
The probe is trained with cross-entropy loss on valid depth pixels within the dataset-specific depth range.
At inference time, continuous depth is recovered as the expected depth under the predicted bin distribution.
We evaluate on ScanNet~\cite{dai2017scannet} and NYUv2~\cite{silberman2012indoor_nyuv2}, training on the ScanNet 25k-frame subset and evaluating on the official ScanNet test scenes, and using the standard NYUv2 train/test split.
All methods use the same probe capacity, optimizer settings, depth-bin configuration, and 50k-iteration training schedule.
We report AbsRel, RMSE, and $\delta_1$, where $\delta_1$ is the percentage of valid pixels satisfying $\max(\hat{d}/d,d/\hat{d})<1.25$.

\subsection{3D Lifting and Rendering}
\label{sec:supp_protocol_3d_lifting}

\paragraph{3D feature uplifting and rendered matching.}
We evaluate whether the learned 2D features remain reliable after explicit lifting into 3D and rendering to novel views.
We use LUDVIG~\cite{marrie2025ludvig}, a learning-free feature uplifting method for Gaussian Splatting scenes.
Given a pretrained Gaussian scene, LUDVIG assigns 2D image features to 3D Gaussians by aggregating features according to each pixel's rendering contribution.
This contribution-weighted procedure avoids an additional feature-optimization stage, allowing us to evaluate the liftability of the frozen 2D representation itself.
We use only the direct uplifting step and do not apply the graph-diffusion refinement from LUDVIG.

We evaluate on Mip-NeRF 360 scenes~\cite{barron2022mipnerf} using Gaussian Splatting models trained for 30,000 iterations with the standard setup~\cite{kerbl20233dgs}.
Every eighth image is held out as a test view; the remaining images are used to train the Gaussian scenes and uplift features.
Following the official LUDVIG implementation, we reduce all features to 40 dimensions with PCA before uplifting to make feature rendering tractable under a common pipeline.
After uplifting, we render the 3D feature field from held-out test poses and evaluate nearest-neighbor correspondence under rendered-to-2D, 2D-to-rendered, and rendered-to-rendered settings.
Matches are evaluated using PCK at the pixel thresholds reported in the main paper.

\paragraph{3D point segmentation.}
We evaluate 3D point segmentation with linear probing on ScanNet~\cite{dai2017scannet} point clouds.
Following the feature aggregation strategy used in OpenScene~\cite{peng2023openscene}, we lift 2D image features onto 3D points using the provided depth maps and camera poses.
For each 3D point, we project it into posed RGB-D frames, sample the corresponding 2D feature at each valid projected pixel, and average the sampled features across valid views.
The aggregated 3D point features are kept frozen, and only a linear classifier is trained for semantic segmentation.
All methods use the same point clouds, camera poses, aggregation procedure, train/validation split, classifier capacity, and optimization settings.
For the fused features, we report mIoU, mAcc, and aAcc.
To expose disagreement that averaging can hide, we additionally apply the same classifier to the feature of every individual view in which a point is visible, without averaging.
We report the mIoU over all point--view pairs (per-view mIoU) and the multi-view agreement, i.e., the probability that two randomly chosen views of a point predict the same class, averaged over all points visible in at least two views.

\end{document}